\documentclass[10pt,journal,compsoc]{IEEEtran}
\usepackage{multirow}
\usepackage{color}
\usepackage{amsmath}
\usepackage{algorithm}
\usepackage{algorithmic}
\usepackage{array}
\usepackage{stfloats}
\usepackage{url}
\usepackage{ragged2e}
\usepackage{cite}

\ifCLASSINFOpdf
  \usepackage[pdftex]{graphicx}
  \graphicspath{{./}}
  \DeclareGraphicsExtensions{.pdf}
\else
  \usepackage[dvips]{graphicx}
  \graphicspath{{./figureeps/}}
  \DeclareGraphicsExtensions{.eps}
\fi

\ifCLASSOPTIONcompsoc
  \usepackage[caption=false,font=footnotesize,labelfont=sf,textfont=sf]{subfig}
\else
  \usepackage[caption=false,font=footnotesize]{subfig}
\fi

\makeatletter
\def\hlinew#1{%
  \noalign{\ifnum0=`}\fi\hrule \@height #1 \futurelet
   \reserved@a\@xhline}
\makeatother%

\definecolor{orange}{RGB}{255,127,0}

\begin{document}
%
\title{Min-Entropy Latent Model for \\Weakly Supervised Object Detection}

\author{Fang Wan,~\IEEEmembership{Student Member,~IEEE},
        Pengxu~Wei,~\IEEEmembership{Student Member,~IEEE},
        Zhenjun~Han,~\IEEEmembership{Member,~IEEE},
        Jianbin~Jiao,~\IEEEmembership{Member,~IEEE},
        and~Qixiang~Ye,~\IEEEmembership{Senior~Member,~IEEE}%

\thanks{F. Wan, Z. Han, J. Jiao, and Q. Ye are with the School of Electronic, Electrical and Communication Engineering, University of Chinese Academy of Sciences (UCAS), Beijing, China, 100049. Emails: {wanfang13@mails.ucas.ac.cn, hanzhj@ucas.ac.cn, jiaojb@ucas.ac.cn, qxye@ucas.ac.cn}. Pengxu Wei is with the School of Data and Computer Science, Sun Yat-sen University, Guangzhou, China. Email: weipengxu11@mails.ucas.ac.cn. }}

\ifCLASSOPTIONpeerreview
\markboth{Journal of \LaTeX\ Class Files,~Vol.~14, No.~8, August~2015}%
{Shell \MakeLowercase{\textit{et al.}}: Bare Demo of IEEEtran.cls for Computer Society Journals}
\fi

\IEEEtitleabstractindextext{%
\begin{abstract}
Weakly supervised object detection is a challenging task when provided with image category supervision but required to learn, at the same time, object locations and object detectors. The inconsistency between the weak supervision and learning objectives introduces significant randomness to object locations and ambiguity to detectors. In this paper, a min-entropy latent model (MELM) is proposed for weakly supervised object detection. Min-entropy serves as a model to learn object locations and a metric to measure the randomness of object localization during learning. It aims to principally reduce the variance of learned instances and alleviate the ambiguity of detectors. MELM is decomposed into three components including proposal clique partition, object clique discovery, and object localization. MELM is optimized with a recurrent learning algorithm, which leverages continuation optimization to solve the challenging non-convexity problem. Experiments demonstrate that MELM significantly improves the performance of weakly supervised object detection, weakly supervised object localization, and image classification, against the state-of-the-art approaches.
\end{abstract}
}

\maketitle

\IEEEdisplaynontitleabstractindextext

\IEEEpeerreviewmaketitle

\IEEEraisesectionheading{\section{Introduction}\label{sec:introduction}}

\IEEEPARstart{S}{{\color{black}upervised}} object detection has made great progress in recent years \cite{girshick2014RCNN, girshick2015fast-rcnn, ren2015faster-rcnn, liu2016ssd, redmon2016yolo, liu2018bow}, as concluded in the object detection survey \cite{liu2018DetectionSurvey}. This can be attributed to the availability of large datasets with precise object annotations and deep neural networks capable of absorbing the annotation information, especially. Nevertheless, annotating a bounding-box for each object in large datasets is laborious, expensive, or even impractical. It is also not consistent with cognitive learning, which requires solely the presence or absence of a class of objects in a scene, instead of bounding-boxes that indicate the precise locations of all objects.

Weakly supervised learning (WSL) refers to methods that rely on training data with incomplete annotations to learn recognition models. Weakly supervised object detection (WSOD) requires solely the image-level annotations indicating the presence or absence of a class of objects in images to learn detectors \cite{andrews2002MIL, pandey2011scene, cinbis2014multi-fold-C, cinbis2015multi-fold-J, bilen2014, song2014pattern, bilen2015, wang2014LCL, wang2015LCL, Song2014On, deselaers2012Loc, Zhang2010Weakly, Siva2011Weakly, Hoffman2015Detector, Bilen2016Weakly, Li2016Weakly, Kantorov2016ContextLocNet, Ren2016Weakly, Ye2017SelLearning, Tang2017OICR, Diba2017WCCN, Jie2017deepself}. It can leverage rich Web images with tags to learn object-level models.

To tackle the WSOD problem, existing approaches often resort to latent variable learning or multi-instance learning (MIL) by using redundant object proposals as inputs. The learning objective is designed to choose a true instance from redundant object proposals of each image to minimize the image classification loss. Due to the unavailability of object-level annotations, WSOD approaches require to collect instances from redundant proposals, as well as learning detectors that compromise the appearance of various objects. It typically requires solving a non-convex model and thus is challenged by the local minimum problem.

\begin{figure}[!t]
    \begin{center}
      \includegraphics[width=1.0\linewidth]{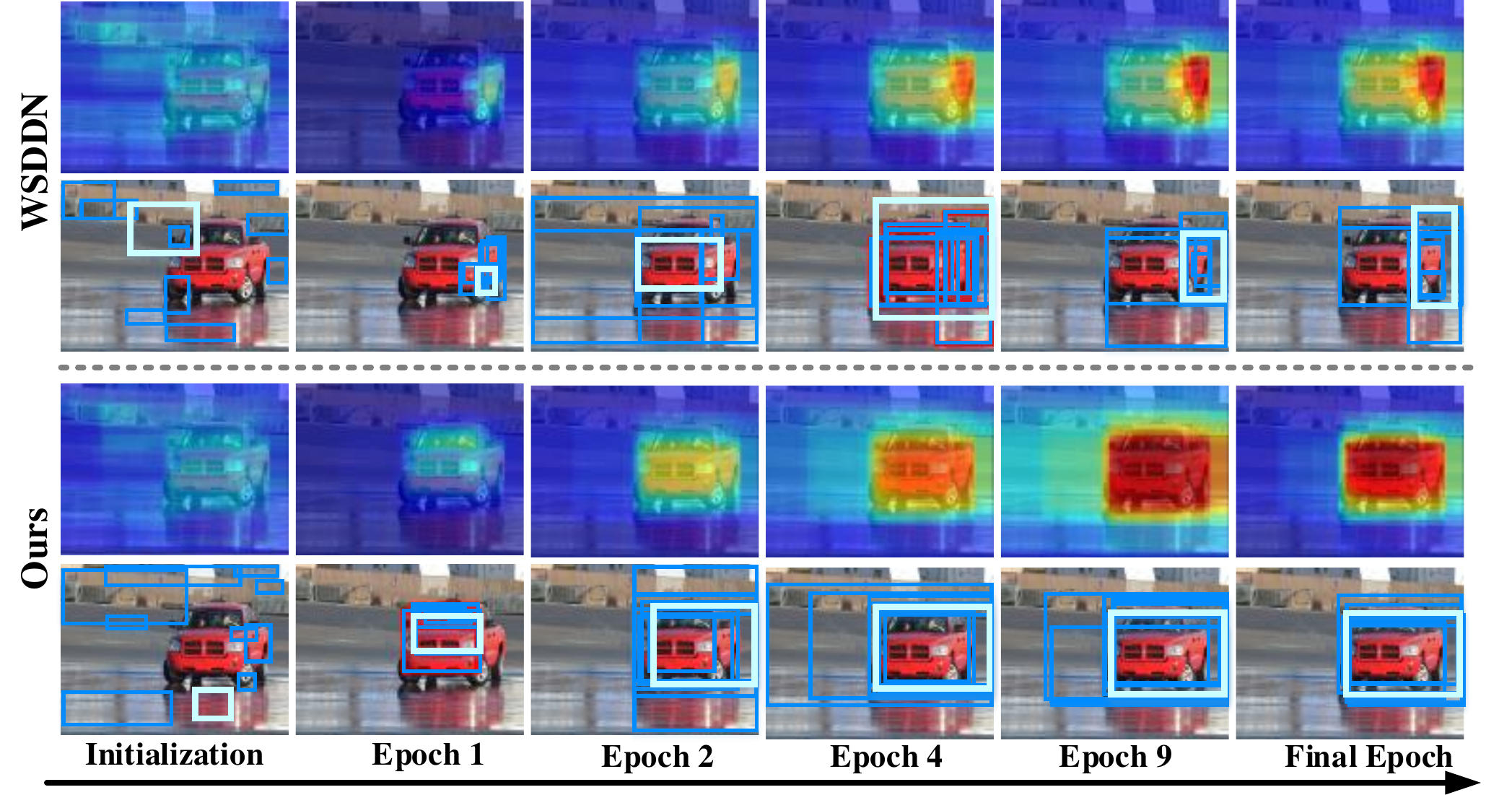}
    \end{center}
    \caption{ Evolution of object locations during learning. Blue boxes denote proposals of high object probability and white ones detected objects. It can be seen that our approach reduces localization randomness and learns object extent. (Best viewed in color.)}
    \label{fig-motivation}
\end{figure}

In the learning procedure of weakly supervised deep detection networks (WSDDN) \cite{Bilen2016Weakly}, a representative WSOD approach, the problem has been observed, $i.e.$, the collected instances switch among different object parts with great randomness, Fig.\ \ref{fig-motivation}.
Various object parts were capable of minimizing image classification loss, but experienced difficulty in optimizing object detectors due to their appearance ambiguity. Recent approaches have used image segmentation \cite{li2016image,Diba2017WCCN}, context information \cite{Kantorov2016ContextLocNet}, and instance classifier refinement \cite{Tang2017OICR} to empirically regularize the learning procedure. However, the issues about principally reducing localization randomness and alleviating the local minimum remain unresolved.

\begin{figure*}[!t]
    \begin{center}
        \includegraphics[width=1\linewidth]{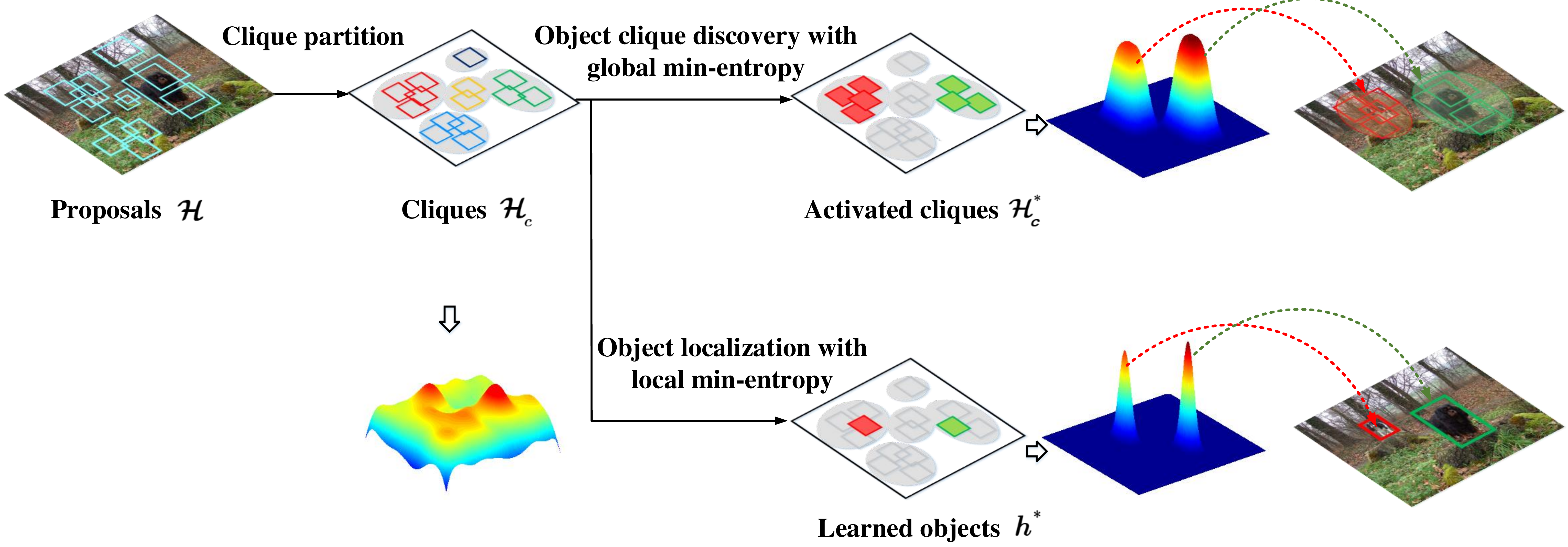}
    \end{center}
    \caption{Illustration of the min-entropy latent model (MELM). A clique partition module is proposed to collect objects/parts from redundant proposals; Based on the cliques, a global min-entropy model is defined for object clique discovery; Within discovered cliques, a local min-entropy model is proposed to suppress object parts and select true objects. The three components are iteratively performed.}
    \label{fig-melm-illustration}
\end{figure*}

\begin{figure}[!t]
    \begin{center}
       \includegraphics[width=1.0\linewidth]{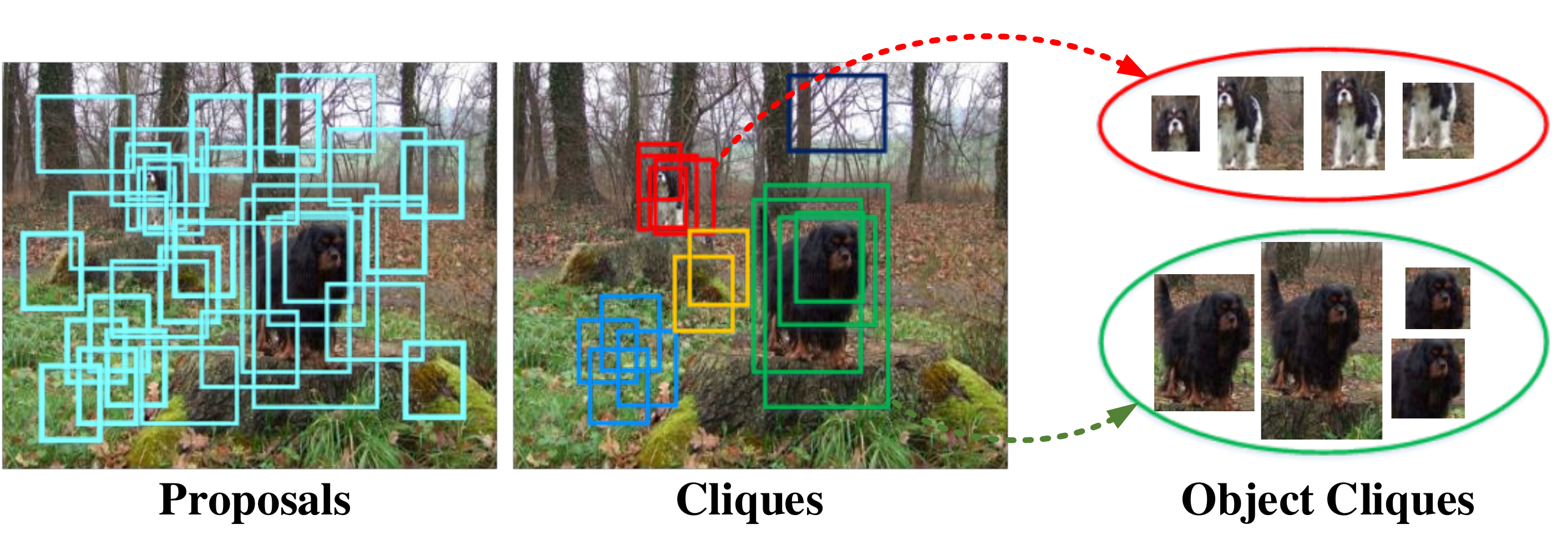}
    \end{center}
    \caption{ The proposals of high scores are selected and dynamically partitioned into same cliques if they are spatially related ($i.e.,$ overlapping with each other) and class related ($i.e.,$ having similar object class scores). Clique partition targets at collecting object/object parts and activating true object extent.}
    \label{clique_partition}
\end{figure}

In this paper, we propose a clique-based min-entropy latent model (MELM) \footnote{Source code is available at https://github.com/WinFrand/MELM.} to collect instances with minimum randomness, motivated by a classical thermodynamic principle: \textit{Minimizing entropy results in minimum randomness of a system.} Min-entropy is used as a model to learn object locations and a metric to measure the randomness of localization during learning. MELM is concluded as three components: (1) Instance (object and object part) collection with a clique partition module; (2) Object clique discovery with a global min-entropy model; (3) Object localization with a local min-entropy model, Fig.\ \ref{fig-melm-illustration}. A clique is defined as a set of object proposals which are spatially related ($i.e.,$ overlapping with each other) and class related ($i.e.,$ having similar object class scores), Fig.\ \ref{clique_partition}. The introduction of proposal cliques can facilitate reducing the redundancy of region proposals and optimizing min-entropy models.

With the clique partition module and min-entropy models, we can collect instances with minimum randomness, activate true object extent, and suppress object parts, Fig.\ \ref{fig-motivation}. MELM is deployed as a clique partition module and network branches concerning object clique discovery and object localization on top of a deep convolutional neural network (CNN). Based on the global and local min-entropy models, we adopt a recurrent strategy to train detectors and pursue true object extent using solely image-level supervision. This is based on the priori that in deep networks the image classification task and object detection task are highly correlated, which allows MELM to recurrently transfer the weak supervision, $i.e.$, image category annotations, to object locations. By accumulating multiple iterations, MELM discovers multiple objects, if such exist, from a single image.

MELM is first proposed in our CVPR paper \cite{wan2018melm} and is promoted both theoretically and experimentally in this full version. The contributions of this paper include: (1) A min-entropy latent model that is integrated with deep networks to effectively collect instances and principally minimize the localization randomness during weakly supervised learning. (2) A clique partition module that facilitates instance collection, object extent activation, and object part suppression. (3) A recurrent learning algorithm that formulates image classification and object detection as a predictor and a corrector, respectively, and leverages continuation optimization to solve the challenging non-convexity problem. (4) State-of-the-art performance of weakly supervised detection,  localization, and image classification.

The remainder of this paper can be concluded as follows. Related works are described in Section 2 and the proposed method is presented in Section 3. Experimental results are given in Section 4. We conclude this paper in Section 5.

\section{Related Work}

WSOD was often solved with a pipelined approach, $i.e.$, an image was first decomposed into object proposals, with which clustering \cite{bilen2015,wang2014LCL,wang2015LCL}, latent variable learning \cite{bilen2014, bilen2015, Song2014On, song2014pattern, wang2014LCL} or multiple instance learning \cite{andrews2002MIL, cinbis2014multi-fold-C, cinbis2015multi-fold-J, Hoffman2015Detector, wang2015relaxed} was used to perform proposal selection and classifier estimation. With the rise of deep learning, pipelined approaches have been evolving into multiple instance learning (MIL) networks \cite{wu2015deep, oquab2015object, shi2016weakly, hoiem2012diagnosing, Bilen2016Weakly, Ren2016Weakly, Li2016Weakly, Kantorov2016ContextLocNet, Jie2017deepself, Diba2017WCCN, Tang2017OICR, zhu2017soft, shi2017weakly}.

\textbf{Clustering.}  Various clustering methods were based on a hypothesis that a class of object instances shape a single compact cluster while the negative instances form multiple diffuse clusters. With such a hypothesis, Wang $et\ al.$ \cite{wang2014LCL,wang2015LCL} calculated clusters of object proposals using probabilistic latent Semantic Analysis (pLSA) on positive samples, and employed a voting strategy on these clusters to determine positive sub-categories. Bilen and Song \cite{song2014pattern, bilen2015} leveraged clustering to initialize latent variables, $i.e.,$ object regions, part configurations and sub-categories, and learn object detectors based on the initialization. Clustering is a simple but effective method.
The disadvantage lies in that a true positive cluster could incorporate significant noise if the objects are surrounded by clutter backgrounds.

\textbf{Latent Variable Learning. } Latent SVM \cite{Ye2017SelLearning} learned object locations and detectors using an Expectation-Maximization-like algorithm. Probabilistic Latent Semantic Analysis \cite{wang2014LCL, wang2015LCL} learned object locations in a latent space.

Various latent variable methods were required to solve the non-convexity problem. They often got stuck in a poor local minimum during learning, $e.g.$, falsely localizing object parts or backgrounds. To pursue a stronger minimum,  object symmetry and class mutual exclusion information \cite{bilen2014},  Nesterov's smoothing \cite{Song2014On}, and convex clustering \cite{bilen2015} were introduced to the optimization function. These approaches can be regarded as regularization which enforces the appearance similarity among objects.

\textbf{Multiple Instance Learning (MIL).} A major approach for tackling WSOD is to formulate it as an MIL problem \cite{andrews2002MIL}, which treats each training image as a ``bag''  and iteratively selects high-scored instances from each bag when learning detectors. However, MIL remains puzzled by random poor solutions. The multi-fold MIL \cite{cinbis2014multi-fold-C, cinbis2015multi-fold-J} used division of a training set and cross validation to reduce the randomness and thereby prevented training from prematurely locking onto erroneous solutions. Hoffman $et\ al.$ \cite{Hoffman2015Detector} trained detectors with weak annotations while transferring representations from extra object classes using full supervision (bounding-box annotation) and joint optimization. To reduce the randomness of positive instances, bag splitting was used during the optimization procedure of MILinear \cite{Ren2016Weakly}.

MIL has been updated to MIL networks \cite{Bilen2016Weakly, Tang2017OICR}, where the convolutional filters behave as detectors to activate regions of interest on the deep feature maps \cite{selvaraju2017grad,kumar2017hide,kim2017two}. The beam search  \cite{bency2016weakly} was used to localize objects  by leveraging spatial distributions and informative patterns captured in the convolutional layers. To alleviate the non-convexity problem, Li $et\ al.$ \cite{Li2016Weakly} adopted progressive optimization as regularized loss functions. Tang $et\ al.$ \cite{Tang2017OICR} proposed to refine instance classifiers online by propagating instance labels to spatially overlapped instances. Diba $et\ al.$ \cite{Diba2017WCCN} proposed weakly supervised cascaded convolutional networks (WCCN). It learned to produce a class activation map
and then selected the best object locations on the map by minimizing the segmentation loss.

MIL networks \cite{Diba2017WCCN, Tang2017OICR, Jie2017deepself} report state-of-the-art performance, but are misled by the inconsistency between data annotations and learning objectives. With image-level annotations, they are capable of learning effective representations for image classification. Without object-level annotation, however, their localization ability is limited. The convolutional filters learned with image-level supervision incorporate redundant patterns, $e.g.$, object parts and backgrounds, which cause localization randomness and model ambiguity.

Recent methods leveraged online instance classifier refinement (OICR) ~\cite{Tang2017OICR, tang2018pcl} and proposal clusters ~\cite{Jie2017deepself, tang2018pcl} to improve localization. The iterative generation of the proposal clusters \cite{tang2018pcl} with OICR prevented the network from concentrating on parts of objects. In this paper, we propose to solve the localization randomness problem by introducing proposal cliques and min-entropy latent models. Our defined proposal cliques facilitate reducing the redundancy of proposals and optimizing min-entropy models. Using the clique-based min-entropy models, we can learn instances with minimum randomness, activate object extent, and suppress object parts, Fig. 1.

To translate the image labels to object locations, the MIL network approaches ~\cite{Tang2017OICR, tang2018pcl} defined multiple network branches: the first one for the basic MIL network and the others for instance classifier refinement. We inherit the multi-branch architecture but add recurrent learning to facilitate the object score feedback \cite{Ke2017}. With recurrent learning, the network branches can directly benefit from each other.

\section{Methodology}

\subsection{Overview}
In weakly supervised learning, the inconsistency between the supervision (image-level annotation) and the objective (object-level classifier) introduces significant randomness to object localization and ambiguity to detectors.
We aim at reducing this randomness to facilitate the collection of instances. To this end, we analyze two factors that cause such randomness: proposal redundancy and location uncertainty. 1) It is known that the objective functions of WSOD models are typically non-convex \cite{andrews2002MIL} and have many local minima. The redundant proposals deteriorate them by introducing more local minima and larger searching space. 2) As the object locations are uncertain, the learned instances may switch among object parts, $i.e.$, local minima.

To reduce the proposal redundancy, we firstly partition the redundant object proposals into cliques and collect instances which are spatially related ($i.e.,$ overlapping with each other) and class related ($i.e.,$ having similar object class scores).  To minimize localization randomness, we design a global min-entropy model that reflects class and spatial distributions of object cliques. By optimizing the global min-entropy model, discriminative cliques containing objects and object parts are discovered, Fig.\ \ref{fig-melm-illustration}, and the cliques which lack discriminative information are suppressed. The discovered cliques are used to activate true object extent.

To localize objects in the discovered cliques, a local min-entropy latent model is defined. By optimizing the local min-entropy model pseudo-objects are estimated and their spatial neighbors are estimated as hard negatives. Such pseudo-objects and hard negatives estimated under the min-entropy principle have minimized randomness during learning, and further improve the performance of object localization, Fig.\ \ref{fig-melm-illustration}. MELM is deployed as a clique partition module and two network branches concerning object clique discovery and object localization, Fig.\ \ref{fig-melm-network}. During learning, it leverages a clique partition module to smooth the objective function and a continuation optimization method to solve the challenging non-convexity problem.

\subsection{Min-Entropy Latent Model}

Let ${x \in {\cal X}}$ denote an image and ${y \in {\cal Y}}$ denote labels indicating if $x$ contains an object or not, where ${{\cal Y}=\{1,0\} }$. $y=1$ indicates that there is at least one object of positive class in the image (positive image) while $y=0$ indicates an image without the object of positive class (negative image).  $h$ denoting an object proposal (location) is a latent variable and ${\cal H}$ denoting object proposals in an image is the solution space.  ${\cal H}_c$ denoting proposal clique is a subset of ${\cal H}$. $\theta$ denotes the network parameters. The min-entropy latent model (MELM) with object locations ${h^*}$ and network parameters $\theta^*$ to be learned, is defined as
\begin{equation}\label{Eq1}
    \begin{split}
    \left\{ {{h^*},{\theta ^*}} \right\}
    &= \mathop {\;\arg \min }\limits_{h,\theta } {E_{\left( {{\cal X,Y}} \right)}}\left( {h,\theta } \right)\\
    &= \mathop {\;\arg \min }\limits_{h,\theta } {E_{\left( {{\cal X,Y}} \right)}}\left( {{{\cal H}_c},\theta } \right) + \lambda {E_{\left( {{\cal X,Y},{{\cal H}_c}} \right)}}\left( {h,\theta } \right)\\
    &\Leftrightarrow \mathop {\arg \min }\limits_{h,\theta } {L_{\left( {{\cal X,Y}} \right)}}\left( {{{\cal H}_c},\theta } \right) + \lambda {L_{\left( {{\cal X,Y},{{\cal H}_c}} \right)}}\left( {h,\theta } \right),
    \end{split}
\end{equation}
where ${E_{\left( {{\cal X,Y}} \right)}}\left( {{{\cal H}_c},\theta } \right)$ and ${E_{\left( {{\cal X,Y},{{\cal H}_c}} \right)}}\left( {h,\theta } \right)$ are the global and local entropy models which respectively serve for object clique discovery and object localization, Fig.\ \ref{fig-melm-network}. $\lambda$ is a regularization weight. ${L_{\left( {{\cal X,Y}} \right)}}\left( {{{\cal H}_c},\theta } \right)$ and ${L_{\left( {{\cal X,Y},{{\cal H}_c}} \right)}}\left( {h,\theta } \right)$ are loss functions based on  ${E_{\left( {{\cal X,Y}} \right)}}\left( {{{\cal H}_c},\theta } \right)$ and ${E_{\left( {{\cal X,Y},{{\cal H}_c}} \right)}}\left( {h,\theta } \right)$, respectively.

Given image-level annotations, $i.e.$, the presence or absence of a class of objects in images, the learning objective of MELM is to find a solution that disentangles object instances from noisy object proposals with minimum image classification loss and localization randomness. To this end, MELM is decomposed into three components including clique partition, object clique discovery, and object localization.

\subsubsection{Clique partition}

Noting that the localization randomness usually occurs among high-scored proposals, we empirically select a set of high-scored (top-200) proposals ${\tilde{\cal H}}$ to construct the cliques, where ${\tilde{\cal H}} \subseteq {\cal H}$.

The proposal cliques are the minimum sufficient cover to ${\tilde{\cal H}}$ which satisfy the following formulations, as
\begin{equation}\label{Eq_clique_partition}
    \left\{ {\begin{array}{*{20}{c}}
{\bigcup\limits_{c = 1}^C {{{\cal H}_c} = {\tilde{\cal H}}} }\\
{\forall c \ne c',\;{{\cal H}_c} \cup {{\cal H}_{c'}} = \emptyset },
\end{array}} \right.
\end{equation}
where $c, c'\in \{1, ..., C\}$ and $C$ is the number of proposal cliques. To partition cliques, the proposals are sorted by their object scores and the following two steps are iteratively performed: 1) Construct a clique using the proposal of highest object score but not belonging to any clique. 2) Find the proposals that overlap with any proposal in the clique larger than a threshold $\tau$ and merge them into the clique.

\subsubsection{Object clique discovery with global min-entropy}
During the learning procedure, it is required that the cliques evolve with minimum randomness. At the same time, it is required to discover discriminative cliques containing objects and object parts. The network parameters fine-tuned with such cliques can activate true object extent. To this end, a global min-entropy model is defined as
\begin{equation}\label{Eq_p_y_H_theta}
\begin{array}{l}
{{\cal H}^{*}_{{c}}} = \mathop {\arg \min }\limits_{{{\cal H}_c}} \;{E_{\left( {{\cal X,Y}} \right)}}\left( {{{\cal H}_c},\theta } \right)\\
\;\;\;\;\;\;\; = \mathop {\arg \min }\limits_{{{\cal H}_c}} \; - \log \sum\limits_c {p\left( {y,{{\cal H}_c};\theta } \right)},
\end{array}
\end{equation}
where $p\left( {y,{{\cal H}_c};\theta } \right)$ is the class probability of a clique ${\cal H}_c$ defined on the object score $s\left( {y,h;\theta } \right)$, as

\begin{equation}\label{Eq4}
    p\left( {y,{{\cal H}_c};\theta } \right){\rm{ = }}\frac{{\exp \left( {1/\left| {{{\cal H}_c}} \right|\sum\limits_{h \in {{\cal H}_c}} {s\left( {y,h;\theta } \right)} } \right)}}{{\sum\limits_{c}\sum\limits_{y} {\exp \left( {1/\left| {{{\cal H}_c}} \right|\sum\limits_{h \in {{\cal H}_c}} {s\left( {y,h;\theta } \right)} } \right)} }},
\end{equation}
where $\left| \cdot \right|$ calculates proposal number in a clique. $s\left(  \cdot  \right)$ denotes the last FC layer in the object clique discovery branch that outputs object scores for proposals.

To ensure that the discovered cliques can best discriminate the positive images from negative ones, we further introduce a classification-related weight $w_{{\cal H}_c}$. Based on the prior that the object class probabilities of proposals are correlated with their image class probabilities, the global min-entropy is then defined as
\begin{equation}\label{Eq_global_entropy}
   {E_{\left( {{\cal X,Y}} \right)}}\left( {{{\cal H}_c},\theta } \right) =  - \log \sum\limits_c {{w_{{{\cal H}_c}}}p\left( {y,{{\cal H}_c};\theta } \right)},
\end{equation}
where $w_{{\cal H}_c}$, defined as
\begin{equation}\label{Eq_w_H_c}
    {w_{{{\cal H}_c}}} = \frac{{p\left( {y,{{\cal H}_c};\theta } \right)}}{{\sum\limits_y {p\left( {y,{{\cal H}_c};\theta } \right)} }},
\end{equation}
is the classification-related weight of clique ${\cal H}_c$. Eq.\ (\ref{Eq_global_entropy}) belongs to the Acz{\'e}l and Dar{\'o}czy (AD) entropy \cite{aczel1963charakterisierung, Entropy2015} family and is derivable. Eq.\ (\ref{Eq_w_H_c}) shows that when $y=1$, ${w_{{{\cal H}_c}}} \in \left[ {0,1} \right]$ is positively correlated to object score of the positive class in a clique, but negatively correlated to scores of all other classes.

\begin{figure*}[!t]
    \begin{center}
       \includegraphics[width=1\linewidth]{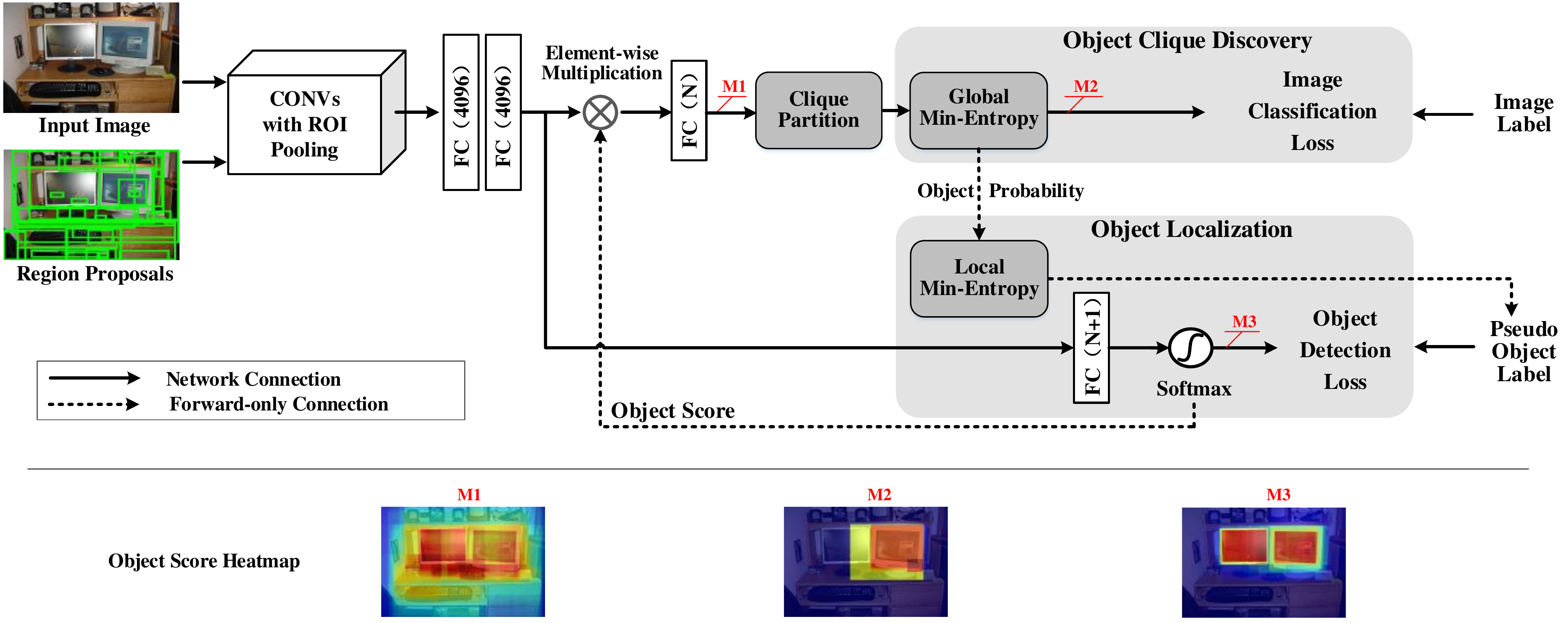}
    \end{center}
    \caption{ {MELM is deployed as a clique partition module and} two network branches for object clique discovery and object localization. These two network branches are unified with  feature learning and optimized with a recurrent learning algorithm. ``M1'', ``M2'' and ``M3'' are heatmaps about proposal scores without min-entropy, with global min-entropy, and with local min-entropy, respectively. $N$ is the number of object categories.}
    \label{fig-melm-network}
\end{figure*}

With above definitions, we implement an object clique discovery branch on top of the network, Fig. \ref{fig-melm-network}, and define a loss function to learn network parameters, as
\begin{equation}\label{loss_clique_discovery}
    \begin{split}
        {L_{\left( {\cal X,Y} \right)}}\left( {{{\cal H}_c},\theta } \right)
        &= y{E_{\left( {\cal X,Y} \right)}}\left( {{{\cal H}_c},\theta } \right) \\
        &\;\;\;- \left( {1 - y} \right)\sum\limits_h {\log \left( {1 - p\left( {y,h;\theta } \right)} \right)}.
    \end{split}
\end{equation}
For positive images, $y = 1$, the second term is zero and only the global min-entropy term is optimized. For negative images, $y = 0$, the first term is zero and the second term (image classification loss) is optimized.

\subsubsection{Object localization with local min-entropy}
The cliques discovered by the global min-entropy model constitute good initialization for object localization, but nonetheless incorporate random false positives, e.g., object parts and/or partial objects with backgrounds. This is caused by the learning objective of object clique discovery, which selects proposals to discriminate positive images from negative ones but does not consider how to precisely localize objects.

A local min-entropy latent model is then defined to localize objects based on the discovered cliques, as
\begin{equation}\label{eq_local_entropy_optim}
    {h^*} = \mathop {\arg \min }\limits_{h \in {{\cal H}_c^*}} {E_{\left( {{\cal X,Y},{{\cal H}_c^*}} \right)}}\left( {h,\theta } \right),
\end{equation}
where
\begin{equation}\label{eq_local_entropy}
    {E_{\left( {{\cal X,Y},{{\cal H}_c}} \right)}}\left( {h,\theta } \right) =  - \sum\limits_{h \in {\Omega _{{h^*}}}} {{w_h}p\left( {y,h;\theta } \right)} \log p\left( {y,h;\theta } \right)
\end{equation}
also belongs to the AD entropy \cite{aczel1963charakterisierung, Entropy2015} family and is also derivable. Different from Eq.\ (\ref{Eq_global_entropy}) which considers the sum of the proposal probabilities globally to predict the image labels, Eq.\ (\ref{eq_local_entropy}) is designed to locally discriminate each proposal to be positive or negative.
$w_h$ is defined as
\begin{equation}\label{eq_w_h}
    {w_h} = \frac{{\sum\limits_{h \in {\Omega _{{h^*}}}} {g\left( {h,{h^*}} \right)} p\left( {y,h;\theta } \right)}}{{p\left( {y,h;\theta } \right)\sum\limits_{h \in {\Omega _{{h^*}}}} {g\left( {h,{h^*}} \right)} }},
\end{equation}
where ${{\Omega _{{h^*}}}}$ denotes neighborhoods of $h^*$ in the clique. $g\left( {h,{h^*}} \right) = {e^{ - a{{\left( {1 - O\left( {h,{h^*}} \right)} \right)}^2}}}$ is a Gaussian kernel function with parameter $a$. $O\left( {h,{h^*}} \right)$ is the IoU of two proposals. The Gaussian kernel function returns a high value when $O\left( {h,{h^*}} \right)$ is large, and a low value when $O\left( {h,{h^*}} \right)$ is small.
With Eq. (\ref{eq_w_h}), we define a ``soft'' proposal labeling strategy for object localization, which is validated to be less sensitive to noises \cite{bagherinezhad2018label} compared to the hard thresholding approach defined in \cite{wan2018melm}.

Accordingly, the loss function of the object localization branch is defined as
\begin{equation}\label{loss_object_localization}
    {L_{\left( {{\cal X,Y},{{\cal H}_c}} \right)}}\left( {h,\theta } \right){\rm{ = }} {E_{\left( {{\cal X,Y},{{\cal H}_c^*}} \right)}}\left( {h,\theta } \right).
\end{equation}
According to the definition of $w_h$, the proposals close to $h^*$ tend to be true objects, and those far from $h^*$, $i.e.$, $O(h; h*) < 0.5$, are hard negatives. Optimizing the loss function produces sparse object proposals of high object probability $p(y,h;\theta)$ and suppresses object parts in clique ${\cal H}_c^*$. During the learning procedure, the localization capability of detectors is progressively improved.

\begin{algorithm}[!ht]
\small
\renewcommand{\algorithmicrequire}{\textbf{Input:}}
\renewcommand\algorithmicensure {\textbf{Output:} }
\caption{\label{Algorithm_1} { Recurrent Learning}}
\begin{algorithmic}[1]
\REQUIRE  {Image $x \in {\cal X}$, image label $y \in {\cal Y}$, and object proposals $h \in {\cal H}$}
\ENSURE  {Network parameters $\theta$}
\STATE  {Initialize object score $s\left( h \right) = s(y,h;\theta) = 1$ for all $h$}
\FOR{ {\ \textbf{$i$} = 1 \textbf{to} $MaxIter$\ }}
\STATE  {$\phi_h \leftarrow$ Compute deep features for all $h$ through forward score }
\STATE  {$\phi_h \leftarrow \phi_h \cdot s(h)$ Aggregate features by object score}
\STATE  {\textbf{Clique partition:}}
\STATE  {\quad ${\cal H}_c \leftarrow$ Clique partition using Eq.\ (\ref{Eq_clique_partition})}
\STATE  {\textbf{\color{black}Object clique discovery:}}
\STATE  {\color{black}\quad ${\cal H}_c^* \leftarrow$ Optimize ${E_{\left( {{\cal X,Y}} \right)}}\left( {{{\cal H}_c},\theta } \right)$ using Eq.\ (\ref{Eq_global_entropy})}
\STATE  {\color{black}\quad ${L_{\left( {\cal X,Y} \right)}}\left( {{{\cal H}_c},\theta } \right) \leftarrow $ Compute using Eq.\ (\ref{loss_clique_discovery}) }
\STATE  {\textbf{Object localization:}}
\STATE  {\color{black}\quad ${h^*} \leftarrow $ Optimize ${E_{\left( {{\cal X,Y},{{\cal H}_c^*}} \right)}}\left( {h,\theta } \right)$ using Eq.\ (\ref{eq_local_entropy_optim})}
\STATE  {\color{black}\quad ${L_{\left( {{\cal X,Y},{{\cal H}_c}} \right)}}\left( {h,\theta } \right) \leftarrow $ Compute using Eq.\ (\ref{loss_object_localization})}
\STATE  {\textbf{Network parameter update:}}
\STATE  {\quad $\theta \leftarrow $ Back-propagate by miniminzing Eq.\ (\ref{loss_clique_discovery}) and Eq.\ (\ref{loss_object_localization})}
\STATE  {$s(h) \leftarrow$ Update object score using parameters $\theta$}
\ENDFOR
\end{algorithmic}
\end{algorithm}

\subsection{Model Implementation}

MELM is implemented with an integrated deep network, with a clique partition module and two network branches added on top of the FC layers, Fig.\ \ref{fig-melm-network}. The first network branch, designated as the \textit{object clique discovery} branch, has a global min-entropy layer, which defines the distribution of object probability and targets at finding candidate object cliques by optimizing the global entropy and the image classification loss. The second branch, designated as the \textit{object localization} branch, has a local min-entropy layer and a soft-max layer. The local min-entropy layer classifies the object candidates in a clique into pseudo objects\footnote{Pseudo objects are the instantaneously learned objects.} and hard negatives by optimizing the local entropy and pseudo object detection loss.

In the learning phase, object proposals are firstly generated for each image. An ROI-pooling layer atop the convolutional layer (CONV5) is used for efficient feature extraction for these proposals. The MELMs are optimized with a recurrently learning algorithm, which uses forward propagation to select sparse proposals as object instances, and back-propagation to optimize the network parameters with the gradient defined in Appendix. The object probability of each proposal is recurrently aggregated by multiplying by the object probability learned in the preceding iteration. In the detection phase, the learned object detectors, $i.e.$, the parameters for the soft-max and FC layers, are used to classify proposals and localize objects.

\subsection{Model Learning}

The objective of model learning is transferring the image category supervision to object locations with min-entropy constraints, $i.e.$, minimum localization randomness.

\begin{figure}[!t]
    \begin{center}
       \subfloat[]{\includegraphics[width=1\linewidth]{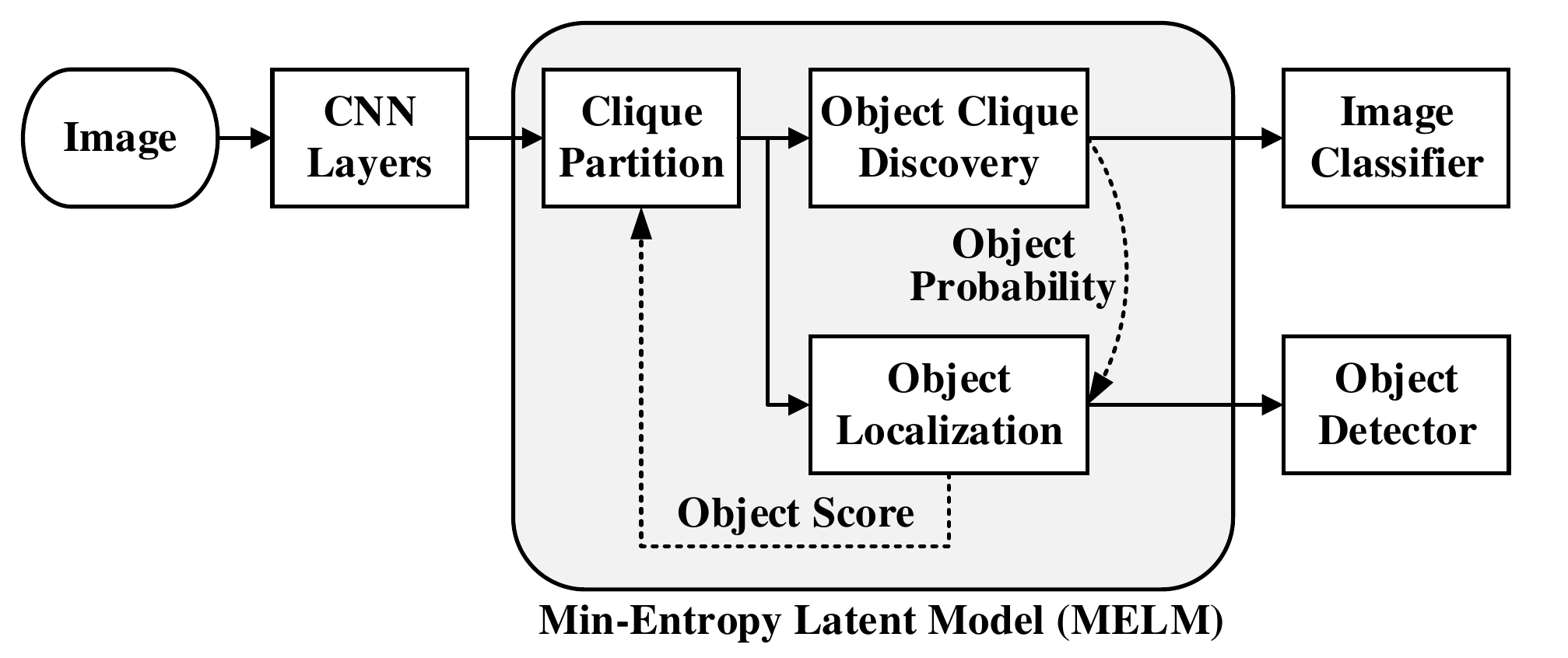}} \\
       \subfloat[]{\includegraphics[width=1\linewidth]{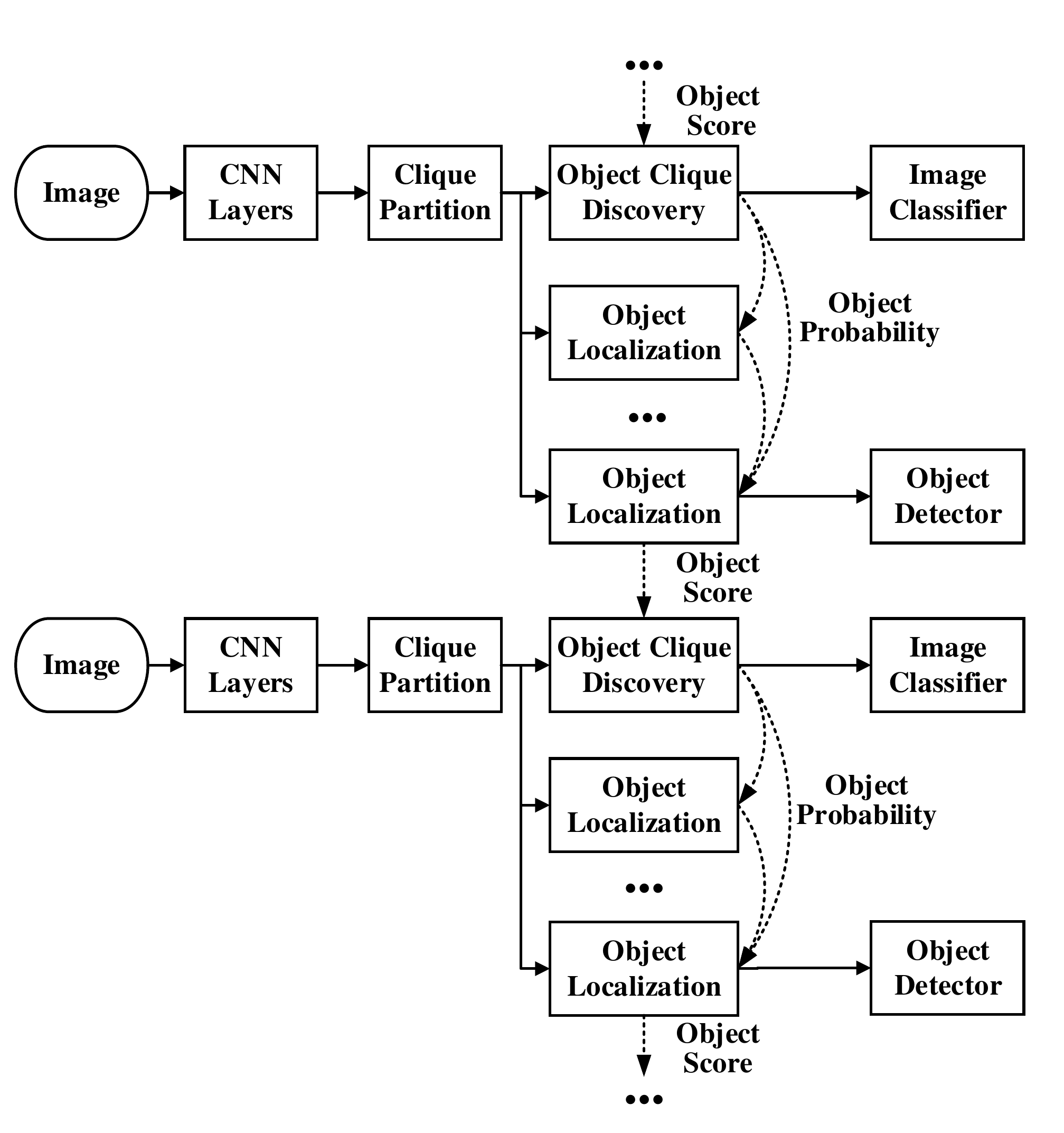}}
    \end{center}
    \caption{Flowchart of (a) the recurrent learning algorithm and (b) unfolded accumulated recurrent learning algorithm. The solid lines denote network connections and dotted lines denote forward-only connections.}
    \label{fig_recurrent}
\end{figure}

\textbf{Recurrent Learning.} A recurrent learning algorithm is implemented to transfer the image-level (weak) supervision using an {\color{black} integrated} forward- and back-propagation procedure, Fig.\ \ref{fig_recurrent}(a). In a feed-forward procedure, the min-entropy latent models discover object cliques and localize objects which are used as pseudo-objects for detector learning. With the learned detectors the object localization branch assigns all proposals new object probability, which is used to aggregate the object scores with an element-wise multiply operation in the next learning iteration. In the back-propagation procedure, the object clique discovery and object localization branches are jointly optimized with an SGD algorithm, which propagates gradients generated with image classification loss and pseudo-object detection loss. With forward- and back-propagation procedures, the network parameters are updated and the image classifiers and object detectors are mutually enforced. The recurrent learning algorithm is described in Alg.\ \ref{Algorithm_1}.

\textbf{Accumulated Recurrent Learning.} Fig.\ \ref{fig_recurrent}(b) shows the proposed accumulated recurrent learning (ARL). In ARL, we add multiple object localization branches, which may localize objects different from those discovered by previous branches. We thus accumulates objects from all previous branches. Doing so not only endows this approach the capability to localize multiple objects in a single image but also improves the robustness about object appearance diversity by learning various objects with multiple detectors.

\subsection{Model analysis}

With the clique partition module and recurrent learning, MELM implements the idea of continuation optimization\cite{Allgower1990Numerical} to alleviate the non-convexity problem.

In continuation optimization, a complex non-convex objective function is denoted as $E(\theta)$, where $\theta$ denotes the model parameters. Optimizing $E(\theta)$ is to find the solution
\begin{equation}\label{eq_continus}
    {\theta^*} = \mathop {\;\arg \min }\limits_{\theta } E(\theta).
\end{equation}
While directly optimizing Eq. (\ref{eq_continus}) causes local minimum solutions, a smoothed function $E(\theta, \lambda)$ is introduced to approximate $E(\theta)$ and facilitate the optimization, as
\begin{equation}
    E(\theta ,\lambda ) = E(\theta ) - \lambda {\cal E}({\theta}),
\end{equation}
where $\lambda \in [0,1]$ controls the smoothness of the approximate function $E(\theta ,\lambda )$ and ${\cal E}({\theta})$ is a correction function. The traditional continuation method traces an implicitly defined curve from a starting point $(\theta^0 , 1)$ to a solution point  $(\theta^* , 0)$, where $\theta^0$ is the solution of $E(\theta, \lambda)$ when $\lambda$=1. During the procedure, if $E(\theta ,\lambda )$ is smooth and its solution is close to $E(\theta )$, we need only to fill the gap between them. This is done by defining a consequence of predictions and corrections to iteratively approximate the original objective function and approach the globally optimal solution $\theta^*$.

The objective function of MELM, defined in Eq. (1), is to find the solution $\{h^*,\theta^*\}$,
\begin{equation}
    \{{h^*},{\theta ^*}\} = \mathop {\;\arg \min }\limits_{h,\theta } {E_{({\cal X},{\cal Y})}}(h,\theta ).
\end{equation}
For the complexity and non-convexity of ${E_{({\cal X},{\cal Y})}}(h,\theta )$, we propose to optimize an approximate function,
\begin{equation}
    {E_{({\cal X},{\cal Y})}}({{\cal H}_c},\theta ) = {E_{({\cal X},{\cal Y})}}(h,\theta ) - \lambda {E_{({\cal X},{\cal Y},{{\cal H}_c})}}(h,\theta ),
\end{equation}
which corresponds to Eq. (1).  ${E_{({\cal X},{\cal Y})}}({{\cal H}_c},\theta )$ is defined by the clique partition module and is smoother than ${E_{({\cal X},{\cal Y})}}(h,\theta )$.
This is achieved by reducing the solution space from thousands of proposals to tens of cliques in each image and averaging the class probability of all proposals in each clique, as defined by Eq.\ (\ref{Eq4}).

With the approximate function defined, we explore recurrent predictions and corrections to optimize the model. The gap between ${E_{({\cal X},{\cal Y})}}({{\cal H}_c},\theta )$ and ${E_{({\cal X},{\cal Y})}}(h,\theta )$ is that the former is defined to discover object cliques but the latter to localize objects. As the solution of ${E_{({\cal X},{\cal Y})}}(h,\theta )$  (object) is included in the solution of ${E_{({\cal X},{\cal Y})}}({{\cal H}_c},\theta )$  (clique), the gap can be simply filled by designing a correction model
${E_{({\cal X},{\cal Y},{{\cal H}_c})}}(h,\theta )$ to localize the object in the clique.
With recurrent learning, the original objective function is thus progressively approximated.

Accordingly, the weakly supervised learning problem is decomposed into an object clique discovery problem (prediction) and object localization problem (correction). The non-convex optimization problem is turned into a proximate problem, which is easier to be optimized \cite{Bengio2009Curriculum, Gulcehre2017Mollifying}.

\subsection{Object Detection}
By optimizing the min-entropy latent models, we obtain object detectors, which are applied to detect objects from test images. The detection procedure involves feature extraction and object localization  Fig.\ \ref{fig-melm-network}. With redundant object proposals extracted by the Selective Search \cite{uijlings2013selective} or the EdgeBox method \cite{Zitnick2014Edge}, a test image is fed to the feature extraction module, and then a ROI-pooling layer is used to extract features for each proposal. The detector outputs object scores for each proposal and a Non-Maximum Suppression (NMS) procedure is used to remove the overlapped proposals.

\section{Experiments}

The PASCAL VOC 2007, 2010, 2012 datasets \cite{everingham2010pascal}, the ILSVRC 2013 dataset \cite{Deng2009ImageNet}, {\color{black}and the MSCOCO 2014 dataset \cite{Lin2014Microsoft}} are used to evaluate the proposed approach. In what follows, the datasets and experimental settings are first described. The evaluation of the model and comparison with the state-of-the-art approaches are then presented.

\subsection{Experimental Settings}

\textbf{Datasets. }
The VOC datasets have 20 object categories. The VOC 2007 datasets contains 9963 images which are divided into three subsets: 5011 for $train$ and $val$, and 4952 for $test$. The VOC 2010 dataset contains 19740 images of which 10103 for $train$ and $val$, and 9637 for $test$. The VOC 2012 dataset contains 22531 images which are divided into three subsets: 11540 for $train$ and $val$, and 10991 for $test$.
The ILSVRC 2013 detection dataset is more challenging for object detection as it has 200 object categories, containing 464278 images where 424126 image for $train$ and $val$, and 40152 images for $test$. For comparison with the previous works, we split the $val$ set of ILSVRC 2013 detection dataset into $val1$ and $val2$ as in \cite{girshick2014RCNN}, which was used for training and test, respectively. Although it has more training images, the number of images for each object category is much less than that in the VOC datasets.
The MSCOCO 2014 dataset contains 80 object categories, with challenging aspects including multiple objects, multiple classes, and small objects.
On the PASCAL VOC and ILSVRC 2013 datasets the mean average precision (mAP) is used for evaluation. On the MSCOCO 2014 dataset the mAP under multiple IoUs is used.

\textbf{CNN Models.} MELM is implemented with two popular CNN models pre-trained on the ImageNet ILSVRC 2012 dataset. The first CNN model VGG-CNN-F (VGGF for short) \cite{chatfield2014return} has a similar architechture as the AlexNet \cite{krizhevsky2012imagenet} which has 5 convolutional layers and 3 fully connected layers. The second CNN model is VGG16 \cite{Simonyan2014Very}, which has 13 convolutional layers and 3 fully connected layers. For these two CNN models, we replaced the spatial pooling layer after the last convolution layer with the ROI-pooling layer as \cite{girshick2015fast-rcnn}. The FC8 layer in the two CNN models was removed and the MELM model was added. 

\textbf{Object Proposals.} The Selective Search \cite{uijlings2013selective} or EdgeBoxes method \cite{Zitnick2014Edge} was used to extract about 2000 object proposals for each image. As the conventional object detection task, we used the fast setting when generating proposals by Selective Search. We also removed the proposals whose width or height are less than 20 pixels.

\textbf{Learning settings.} {\color{black}Following \cite{Bilen2016Weakly,Kantorov2016ContextLocNet,Tang2017OICR,Diba2017WCCN}, the input images
were re-sized into 5 scales \{480, 576, 688, 864, 1200\} with respect to the larger side, height or width. 
The scale of a training image was randomly selected and each image was randomly flipped.} In this way, each test image was augmented into 10 images. For recurrent learning, we employed the SGD algorithm with momentum 0.9, weight decay 5e-4, and batch size 1. The model iterated 20 epochs where the learning rate was 5e-3 for the first 15 epochs and 5e-4 for the last 5 epochs. The output scores of each proposal from the 10 augmented images were averaged.

\subsection{Model Effect and Analysis}

\subsubsection{Clique Affect}

\begin{figure*}[!t]
    \begin{center}
        \subfloat[Cliques and objects]{\includegraphics[width=0.5\linewidth]{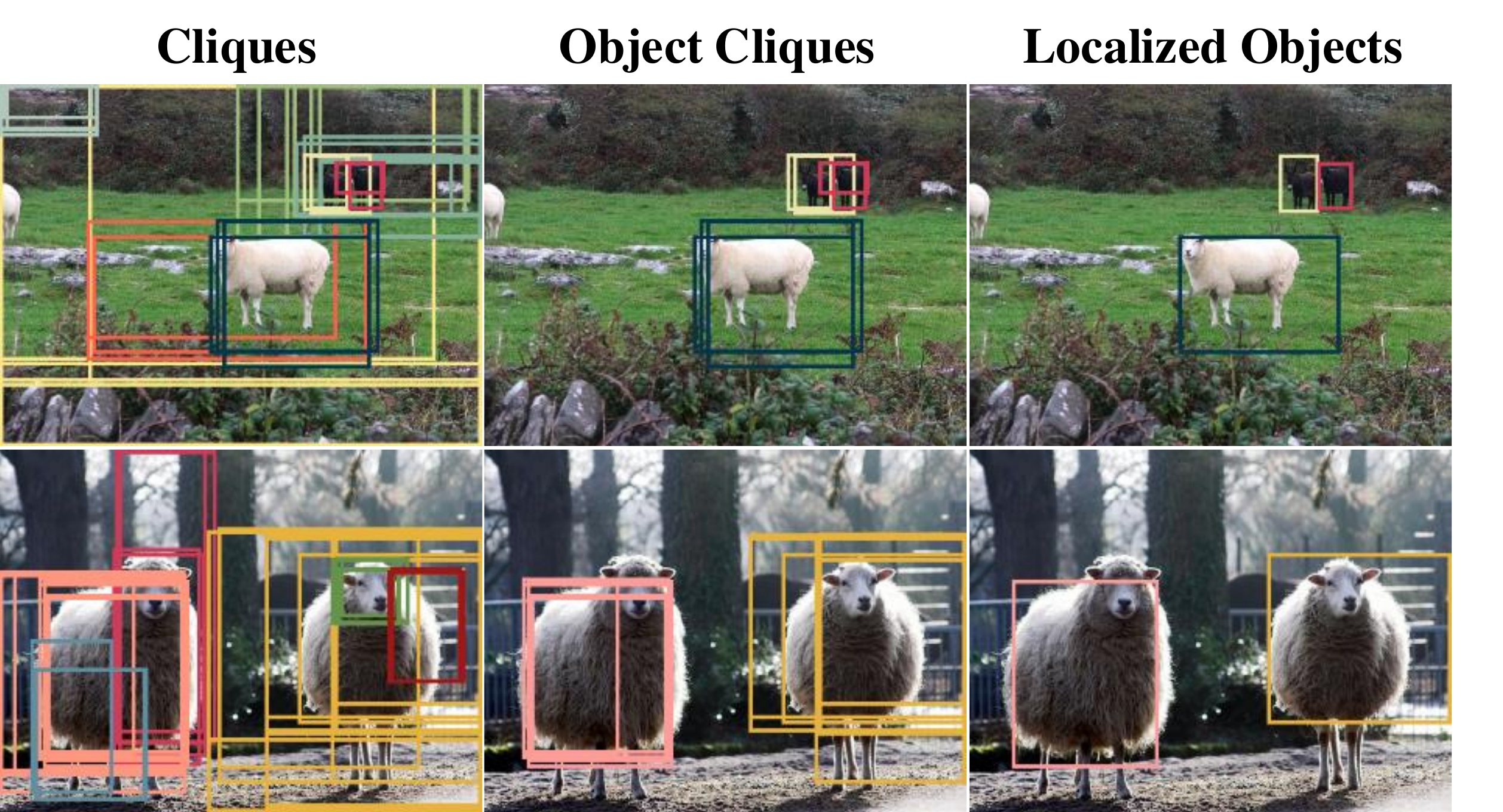}}
        \subfloat[Score maps]{\includegraphics[width=0.5\linewidth]{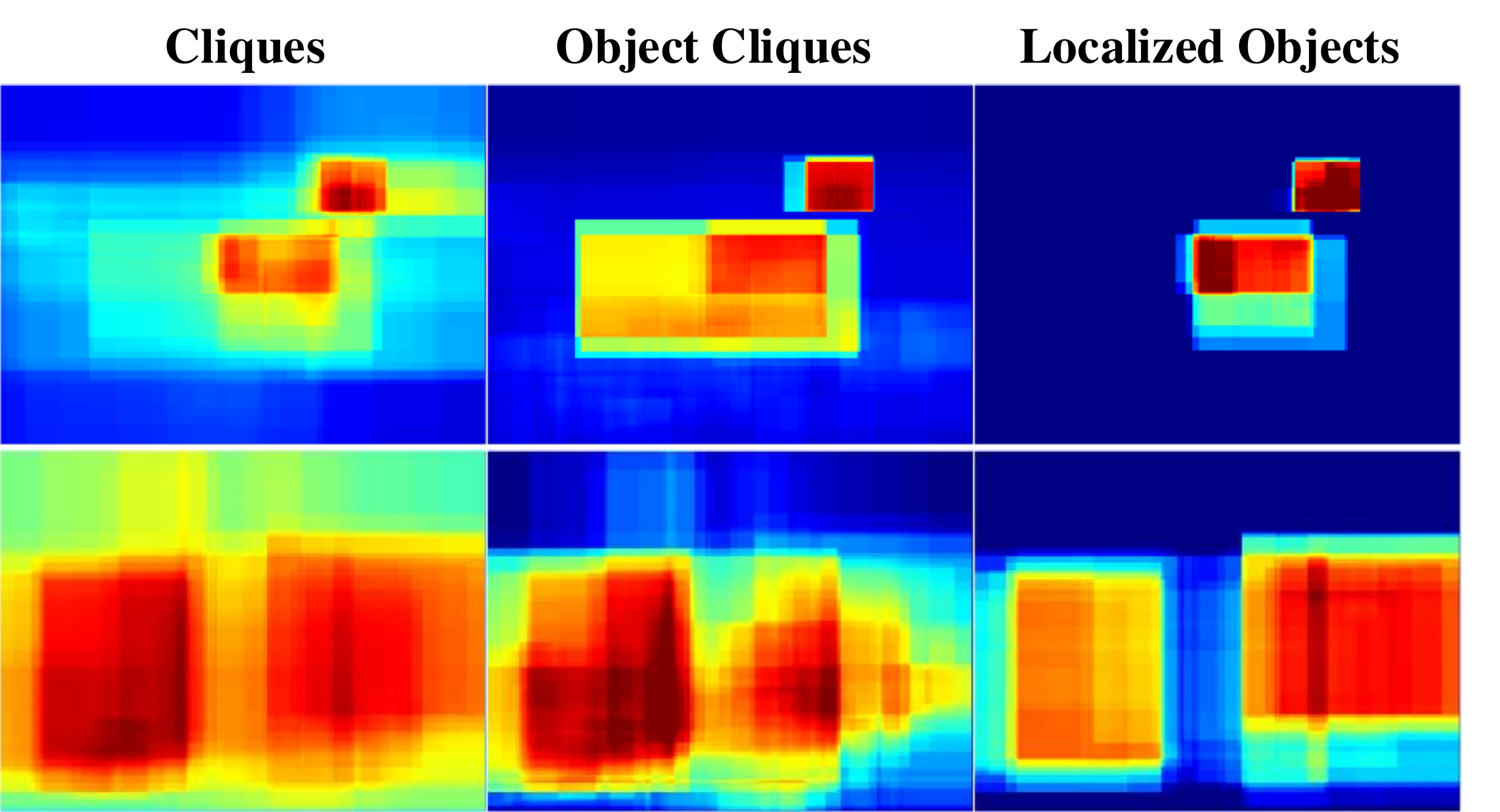}}
    \end{center}
    \caption{ {\color{black} Visualization of the clique partition, object clique discovery, and object localization results. (a) Bounding boxes of different colors denote proposals from different cliques. (b) Score maps of cliques and objects.  (Best viewed in color)
    }}
    \label{fig-clique-results}
\end{figure*}

\begin{figure}[!t]
    \begin{center}
       \includegraphics[width=1\linewidth]{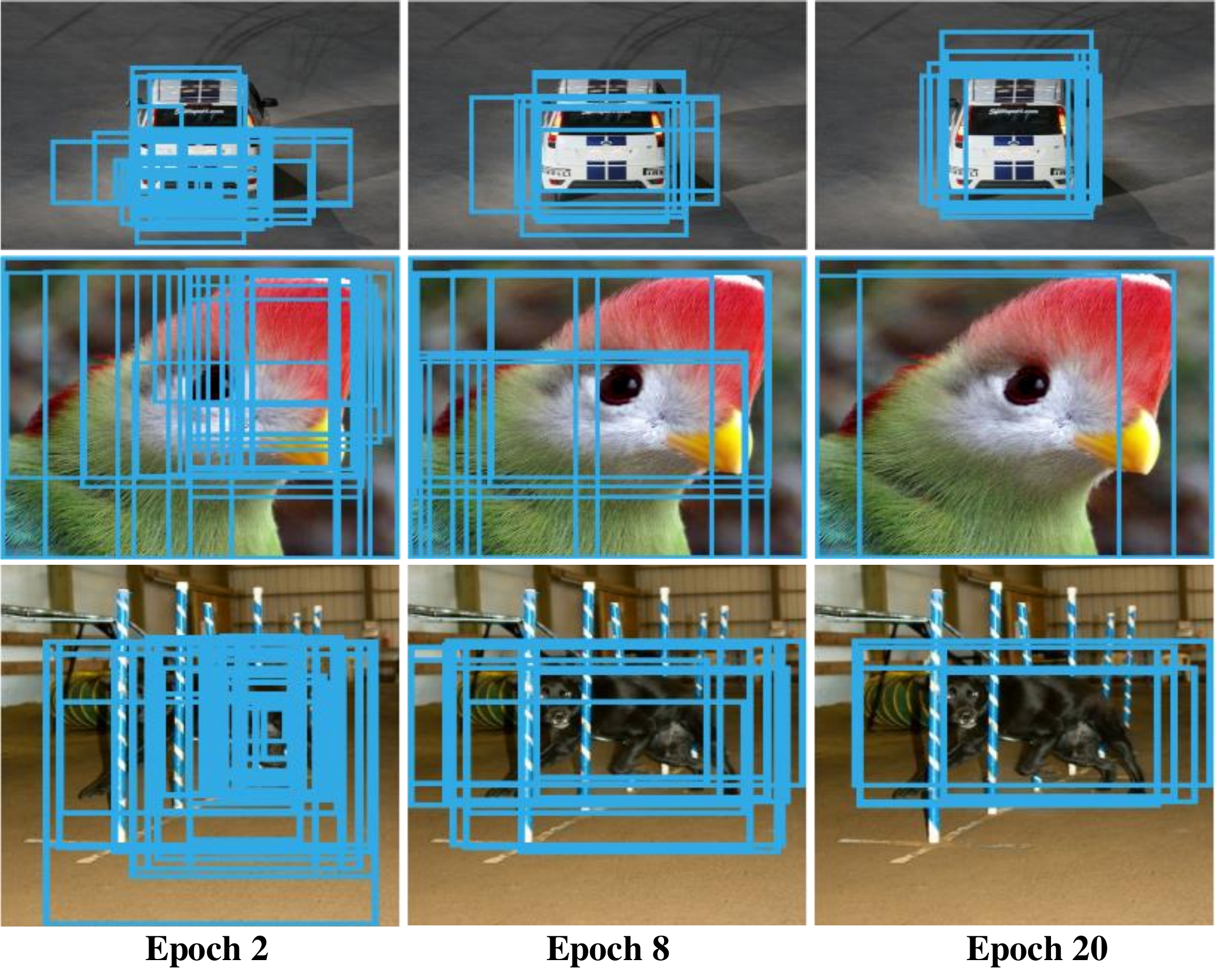}
    \end{center}
    \caption{Evolution of cliques. (Best viewed in color).}
    \label{clique_results}
\end{figure}

{\color{black}
Fig.\ \ref{fig-clique-results} shows that in discovered cliques discriminative objects and object parts were collected and the proposals which lack discriminative information were suppressed.
With the proposals about objects and object parts, the global min-entropy model could activate object extent during the back-propagation procedure. It can also be seen that the true object in a clique can be precisely localized after the recurrent learning procedure.

Fig. \ref{clique_results} shows the object cliques from different learning epochs. It can be seen that in the early training stage (Epoch 2), the object clique collected the object extent, $i.e$, object and object parts. This ensured the object extent activation by the object clique discovery branch. The object localization branch further suppressed the object parts in the object clique (Epoch 4).
MELM finally activated the true object extent, suppressed the object part and detected objects accurately (Epoch 20).
}

\subsubsection{Randomness Analysis}

\begin{figure}[!t]
    \begin{center}
       \subfloat[]{\includegraphics[width=0.5\linewidth]{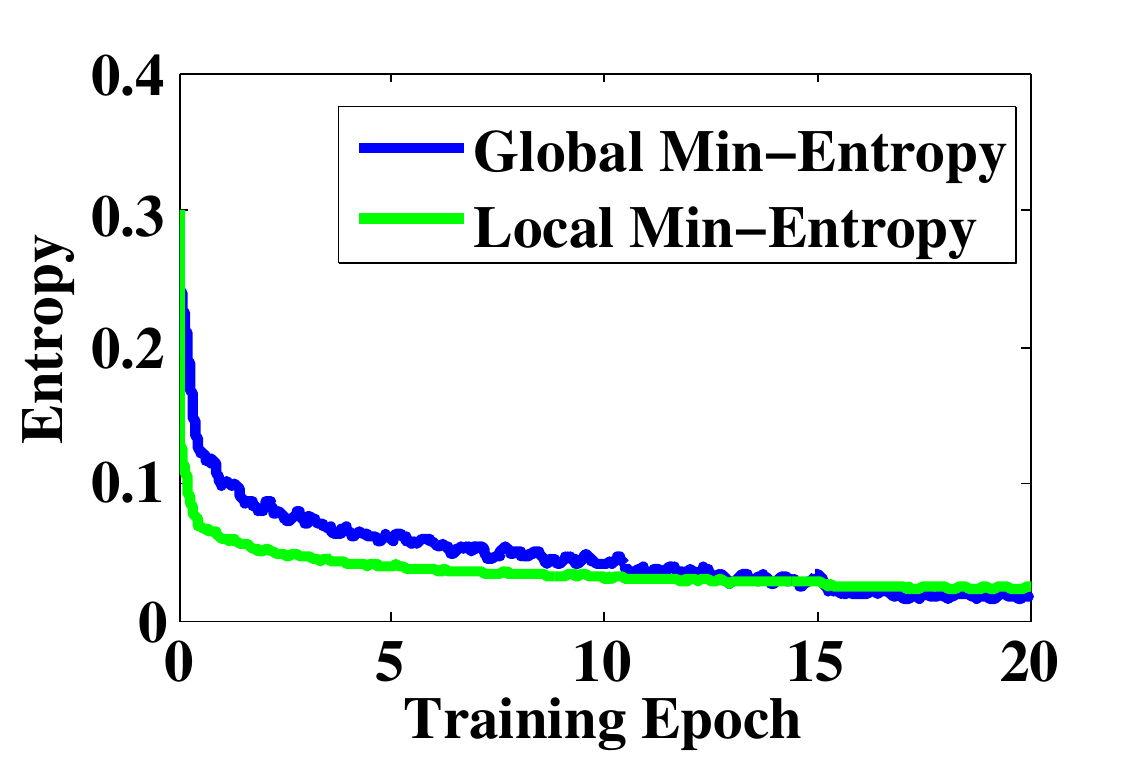}}
       \subfloat[]{\includegraphics[width=0.5\linewidth]{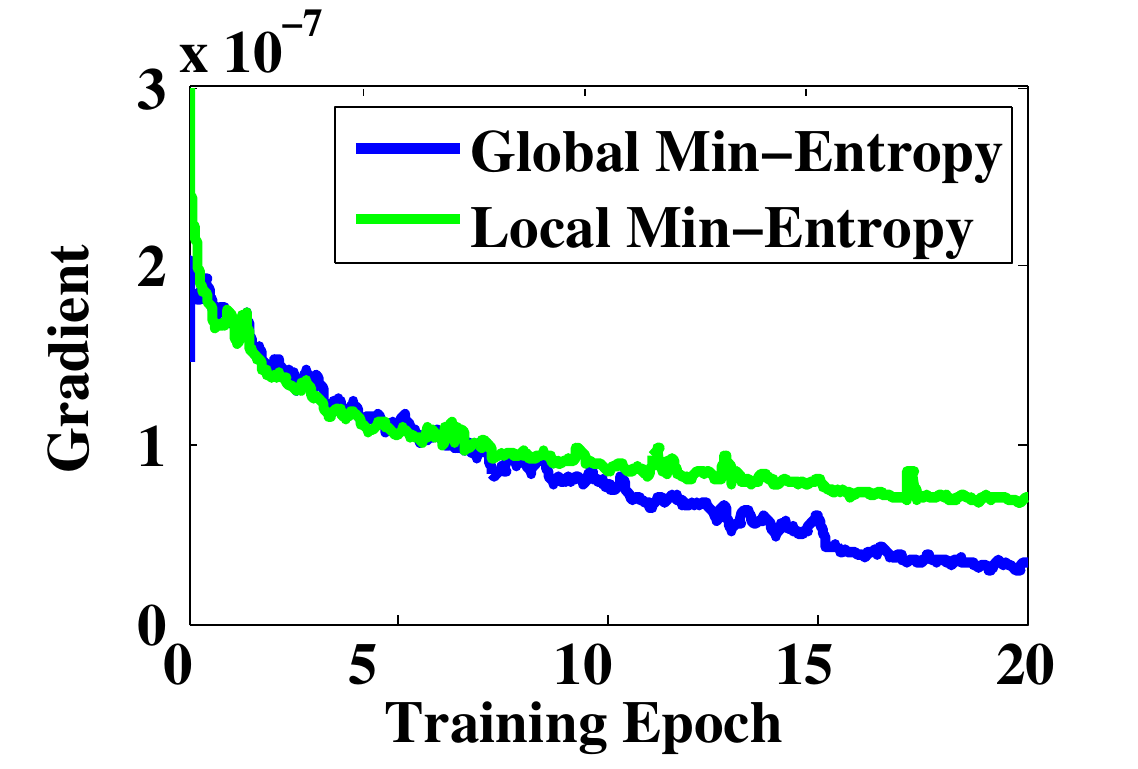}} \\
       \subfloat[]{\includegraphics[width=0.5\linewidth]{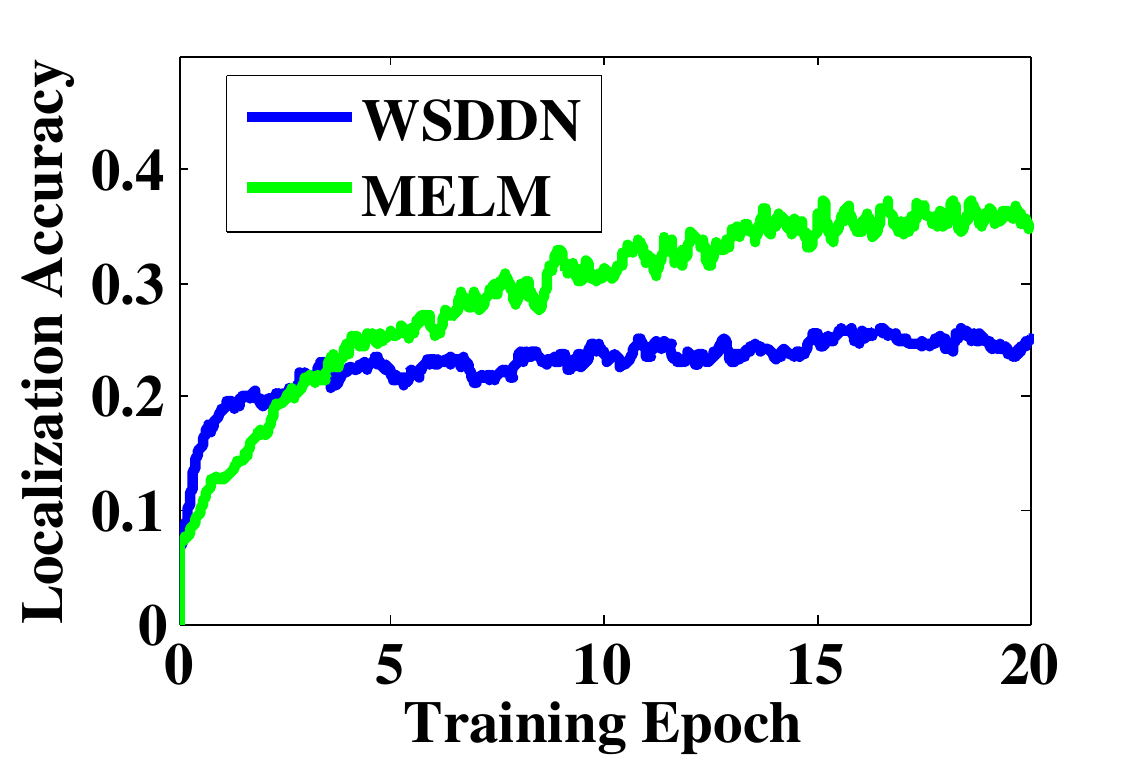}}
       \subfloat[]{\includegraphics[width=0.5\linewidth]{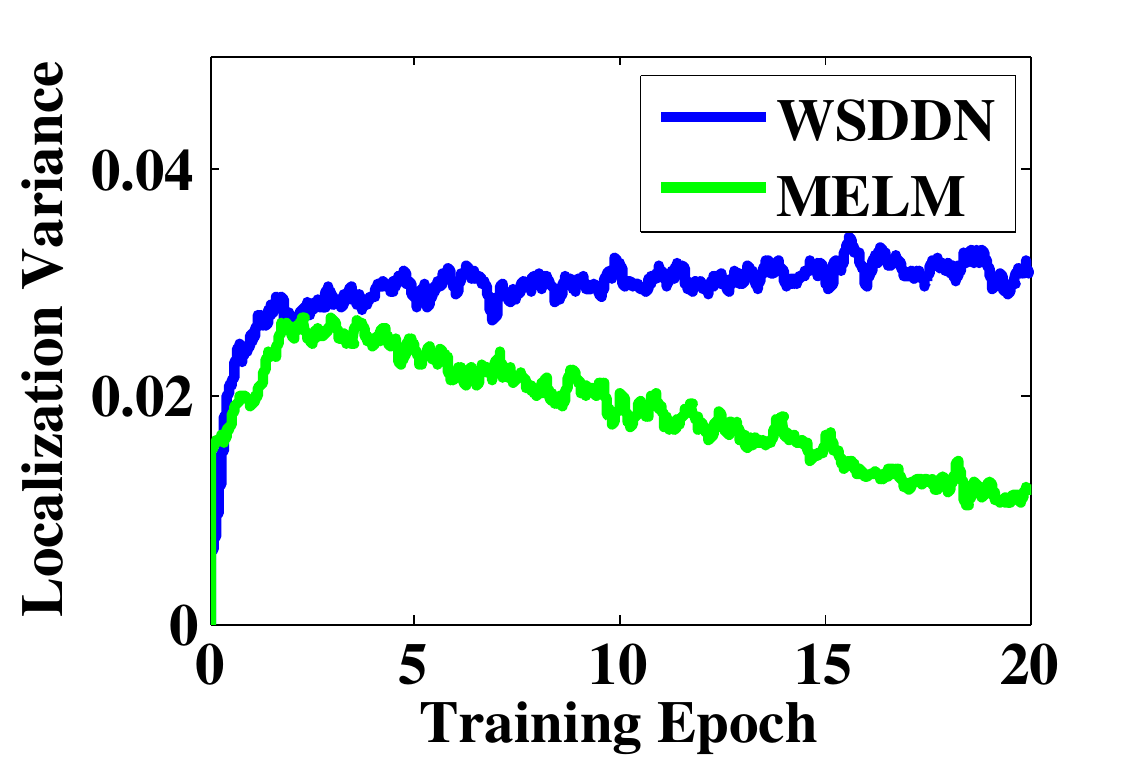}}
    \end{center}
    \caption{ {Gradient, entropy, and localization on the PASCAL VOC 2007 $trainval$ set. (a) The evolution of entropy. (b) The evolution of gradient. (c) Localization accuracy. (d) Localization variance. }}
    \label{entropy_gradient_training}
\end{figure}

Fig.\ \ref{entropy_gradient_training}a shows the evolution of global and local entropy, suggesting that our approach optimizes the min-entropy objective during learning. Fig.\ \ref{entropy_gradient_training}b provides the gradient evolution of the FC layers. In the early learning epochs, the gradient of the global min-entropy module was slightly larger than that of the local min-entropy module, suggesting that the network focused on optimizing the image classifiers. As learning proceeded, the gradient of the global min-entropy module decreased such that the local min-entropy module dominated the training of the network, indicating that the object detectors were being optimized.

To evaluate the effect of min-entropy, the randomness of object locations was evaluated with localization accuracy and localization variance.
Localization accuracy was calculated by weighted averaging the overlaps between the ground-truth object boxes and the learned object boxes, by using $p(y,h;\theta)$ as the weight. 
Localization variance was defined as the weighted variance of the overlaps by using $p(y,h;\theta)$ as the weight. 
Fig.\ \ref{entropy_gradient_training}c and Fig.\ \ref{entropy_gradient_training}d show that the proposed MELM had significantly greater localization accuracy and lower localization variance than WSDDN. This strongly indicates that our approach effectively reduces localization randomness during weakly supervised learning.

\begin{figure*}[!t]
    \begin{center}
       \includegraphics[width=0.800\linewidth]{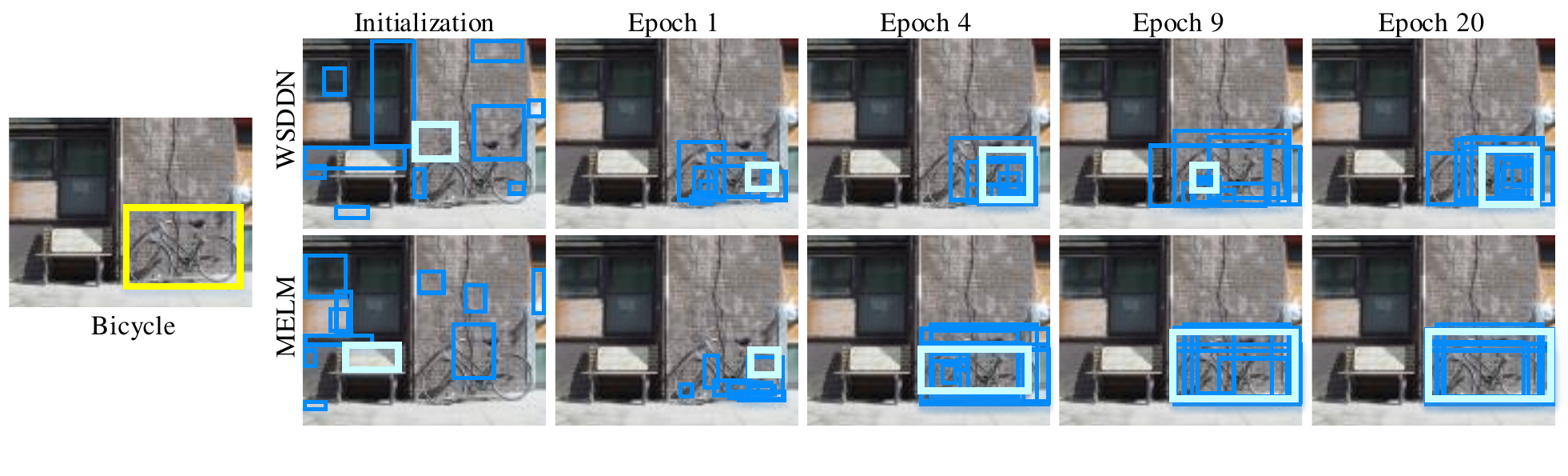}
       \includegraphics[width=0.155\linewidth]{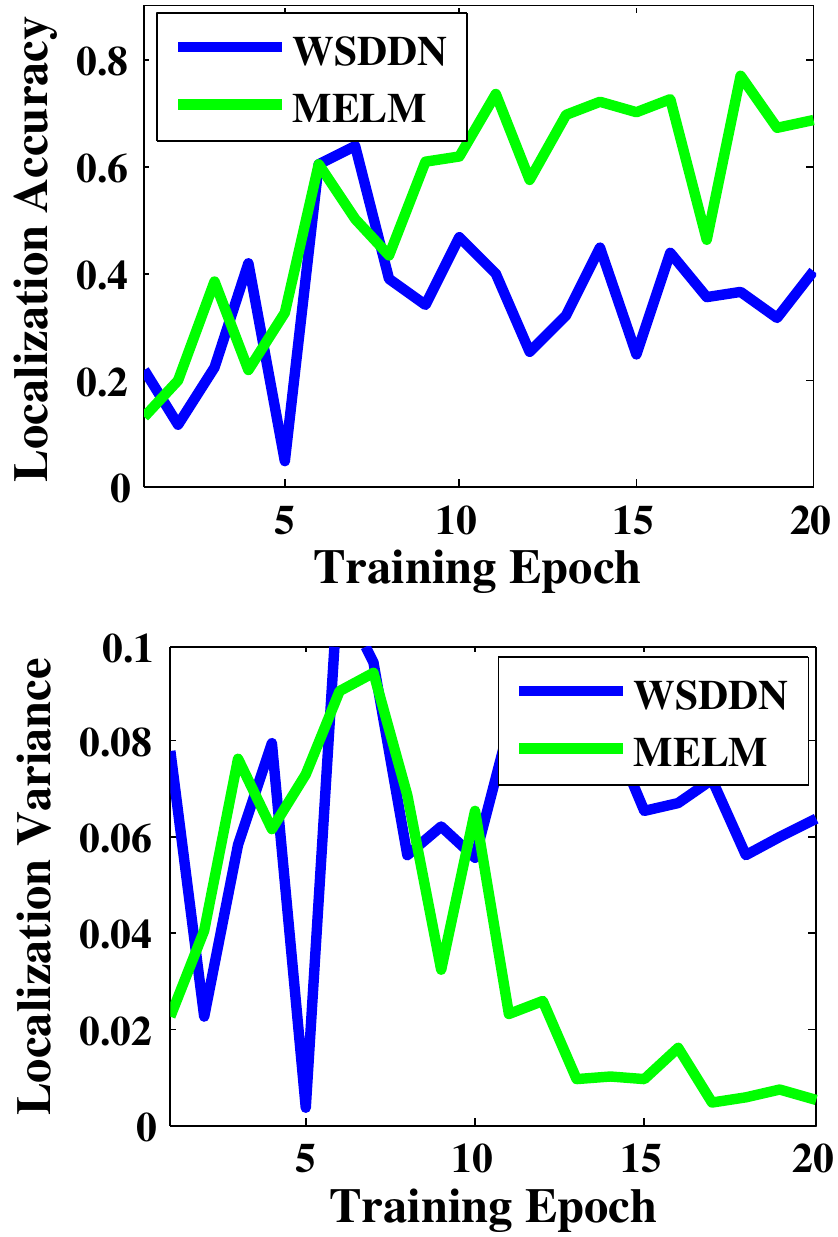}
       \includegraphics[width=0.800\linewidth]{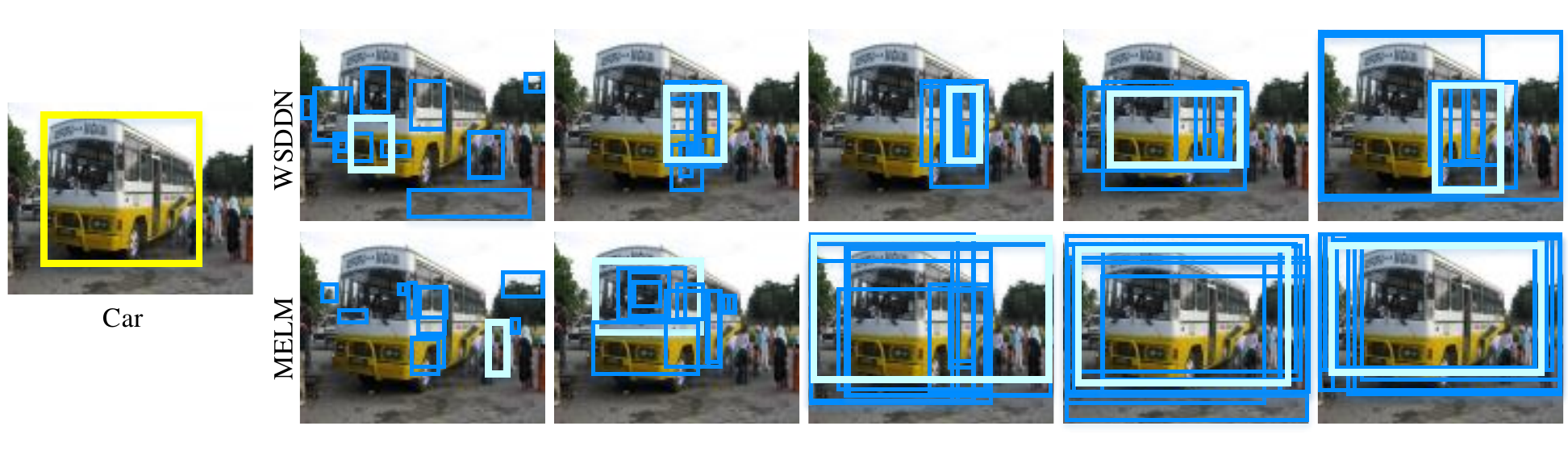}
       \includegraphics[width=0.155\linewidth]{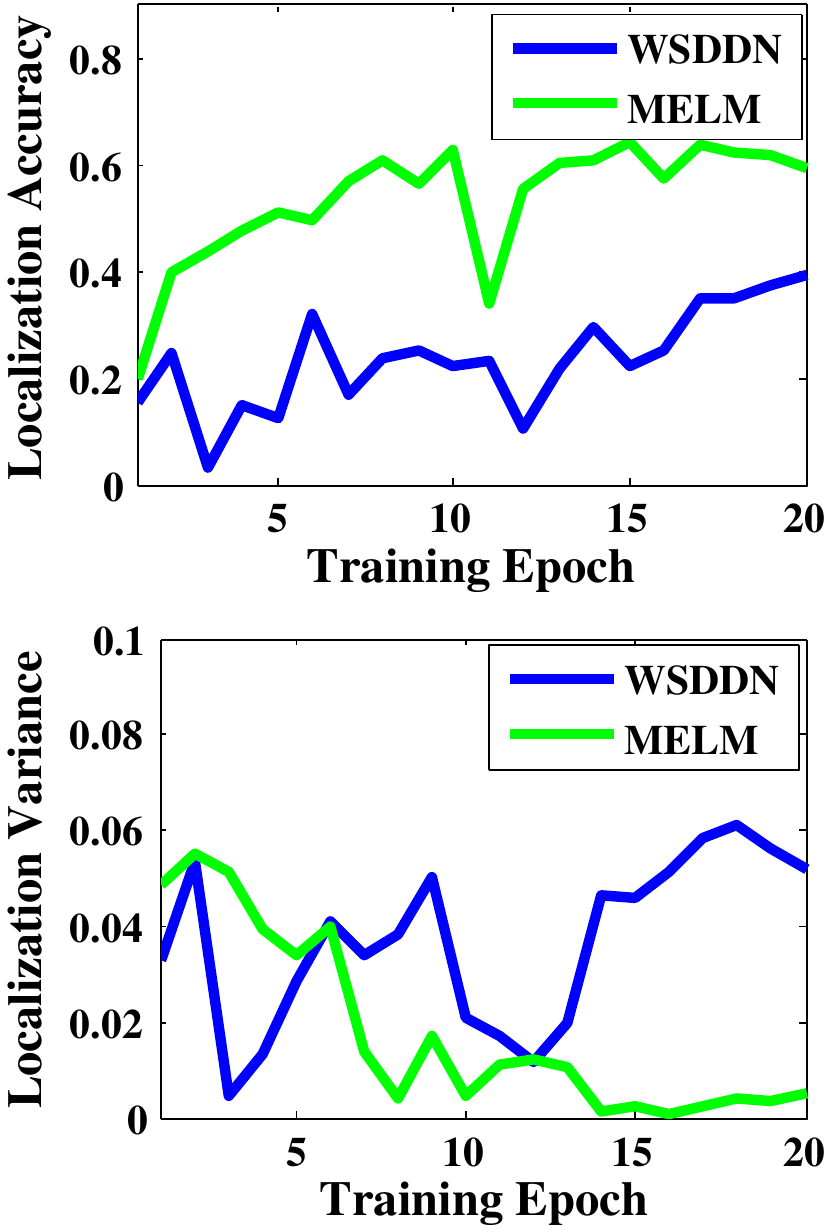}
    \end{center}
    \caption{Comparison of the learned object locations by WSDDN \cite{Bilen2016Weakly} and MELM. The yellow boxes in the first column denote ground-truth objects. The white boxes denote the learned object locations and the blue boxes denote the high-scored proposals. It can be seen that for WSDDN the learned object locations evolved with large randomness, $i.e.$, switch among the proposals around the objects. In contrast the object locations learned by MELM are consistent and have small randomness, which is quantified by the localization variance curves in the last column.  (Best viewed in color)}
    \label{fig-randomness-results}
\end{figure*}

Such an effect was further illustrated in Fig.\ \ref{fig-randomness-results}, where we compared WSDDN with MELM by the localization accuracy and localization variance during the learning.
As shown in Fig. \ \ref{fig-randomness-results}, MELM significantly reduced the localization randomness and achieved higher localization accuracy than WSDDN. Take the ``bicycle'' in Fig. \ \ref{fig-randomness-results} for example.  In the early training epochs, both WSDDN and MELM failed to localize the objects. In the following training epochs MELM reduced the randomness and achieved high localization accuracy. In contrast, WSDDN switched among object parts and failed to localize the true objects.

\begin{table}[!t]
\renewcommand{\arraystretch}{1.3}
\caption{Detection mean average precision (\%) on the PASCAL VOC 2007 test set. Ablation experimental results of MELM.}
\label{table_comp_ourself}
\centering
\scriptsize
\begin{tabular}{@{}p{2.0cm}<{\centering}@{}|p{3cm}<{\centering}|@{}p{1.5cm}<{\centering}@{}}
\hlinew{1.4pt}
CNN & Method & mAP  \\
\hline
\multirow{6}{*}{VGGF } & MELM-base        & 31.5\\
                       & MELM-base+Clique & 33.9 \\
                       & MELM-D           & 33.6 \\
                       & MELM-L           & 36.0 \\
                       & MELM-D+RL        & 34.1 \\
                       & MELM-L+RL        & \textbf{38.4} \\
\hline
\multirow{8}{*}{VGG16} & MELM-base+Clique & 29.5 \\
                       & MELM-D           & 32.6 \\
                       & MELM-L           & 40.1 \\
                       & MELM-D+RL        & 34.5 \\
                       & MELM-L+RL        & 42.6 \\
                       & MELM-D+ARL       & 37.4 \\
                       & MELM-L1+ARL      & 46.4 \\
                       & MELM-L2+ARL      & \textbf{47.3} \\
\hlinew{1.4pt}
\end{tabular}
\end{table}

\subsubsection{Ablation Experiments}

\textbf{Baseline.}
The baseline approach was derived by simplifying Eq.\ (\ref{loss_clique_discovery}) to solely model the global entropy $E_{({\cal X,Y})}({\cal H}_c,\theta)$. This is similar to WSDDN without the spatial regulariser \cite{Bilen2016Weakly} where the single learning objective is to minimize the image classification loss. This baseline, referred to as ``MELM-base'' in Table \ \ref{table_comp_ourself}, achieved 31.5\% mAP using the VGGF network.

\textbf{Clique Effect.}
By dividing the object proposals into cliques, the ``MELM-base'' approach was promoted to ``MELM-base+Clique''. 
Table \ \ref{table_comp_ourself} shows that the introduction of proposal cliques improved the detection performance by 2.4\% (from 31.5\% to 33.9\%). That occurred because using partitioned cliques reduced the solution space of the latent variable learning, thus readily reducing the redundancy of object proposals and facilitating a better solution. We also conducted experiments with different $\tau$ values, which controls the clique size as defined in Sec. 3.2.1, and summarized the results in Table. \ref{table_clique}. {\color{black}Accordingly, we empirically set $\tau$ to be 0.7 in other experiments.}

\begin{table}[!t]
\renewcommand{\arraystretch}{1.3}
\newcommand{\tabincell}[2]{\begin{tabular}{@{}#1@{}}#2\end{tabular}}
\caption{{\color{black}Detection mean average precision (\%) on the PASCAL VOC 2007 $val$ set. Performance with different clique sizes (controlled by $\tau$) of MELM.}}
\label{table_clique}
\centering
\scriptsize
\color{black}
\begin{tabular}{
p{1.7cm}<{\centering}|
p{0.7cm}<{\centering}|
p{0.7cm}<{\centering}|
p{0.7cm}<{\centering}|
p{0.7cm}<{\centering}|
p{0.7cm}<{\centering}|
p{0.7cm}<{\centering}
}
\hlinew{1.4pt}
$\tau$ & 0.1  & 0.3  & 0.5   & 0.7  & 0.9   & 1 \\
\hline
mAP    & 32.6 & 34.3 & 34.4  & 35.3 & 33.5  & 34.4 \\
\hlinew{1.4pt}
\end{tabular}
\end{table}

\begin{table*}[!t]
\renewcommand{\arraystretch}{1.3}
\newcommand{\tabincell}[2]{\begin{tabular}{@{}#1@{}}#2\end{tabular}}
\caption{Detection mean average precision (\%) on the PASCAL VOC 2007 test set. Comparison of MELM to the state-of-the-arts.}
\label{table_comp_state_of_the_art}
\centering
\scriptsize
\begin{tabular}{@{}p{1.0cm}<{\centering}@{}|p{2.5cm}
@{}m{0.65cm}<{\centering}@{}m{0.65cm}<{\centering}@{}m{0.65cm}<{\centering}@{}m{0.65cm}<{\centering}@{}m{0.80cm}<{\centering}
@{}m{0.65cm}<{\centering}@{}m{0.60cm}<{\centering}@{}m{0.60cm}<{\centering}@{}m{0.65cm}<{\centering}@{}m{0.60cm}<{\centering}
@{}m{0.65cm}<{\centering}@{}m{0.65cm}<{\centering}@{}m{0.65cm}<{\centering}@{}m{0.80cm}<{\centering}@{}m{0.80cm}<{\centering}
@{}m{0.75cm}<{\centering}@{}m{0.80cm}<{\centering}@{}m{0.75cm}<{\centering}@{}m{0.65cm}<{\centering}@{}m{0.60cm}<{\centering}
@{}m{0.85cm}<{\centering}@{}}
\hlinew{1.4pt}
CNN & Method & aero & bike & bird & boat & bottle & bus & car & cat & chair & cow & table & dog & horse & mbike & person & plant & sheep & sofa & train & tv & mAP  \\
\hline
\multirow{9}{*}{\tabincell{c}{VGGF/\\AlexNet}}
& MILinear \cite{Ren2016Weakly} &
41.3 & 39.7 & 22.1 &  9.5 &  3.9 & 41.0 & 45.0 & 19.1 &  1.0 & 34.0 &
16.0 & 21.3 & 32.5 & 43.4 & \textbf{21.9}& 19.7 & 21.5 & 22.3 & 36.0 & 18.0 & 25.4 \\
& Multi-fold MIL \cite{cinbis2015multi-fold-J} &
39.3 & 43.0 & 28.8 & 20.4 &  8.0 & 45.5 & 47.9 & 22.1 &  8.4 & 33.5 &
23.6 & 29.2 & 38.5 & 47.9 & 20.3 & 20.0 & 35.8 & 30.8 & 41.0 & 20.1 & 30.2 \\
& PDA \cite{Li2016Weakly} &
49.7 & 33.6 & 30.8 & 19.9 & 13.0 & 40.5 & 54.3 & 37.4 & \textbf{14.8} & 39.8 &
9.4  & 28.8 & 38.1 & 49.8 & 14.5 & \textbf{24.0} & 27.1 & 12.1 & 42.3 & 39.7 & 31.0 \\
& LCL+Context \cite{wang2015LCL} &
48.9 & 42.3 & 26.1 & 11.3 & 11.9 & 41.3 & 40.9 & 34.7 & 10.8 & 34.7 &
18.8 & 34.4 & 35.4 & 52.7 & 19.1 & 17.4 & 35.9 & 33.3 & 34.8 & 46.5 & 31.6 \\
& WSDDN \cite{Bilen2016Weakly} &
42.9 & 56.0 & 32.0 & 17.6 & 10.2 & 61.8 & 50.2 & 29.0 &  3.8 & 36.2 &
18.5 & 31.1 & 45.8 & 54.5 & 10.2 & 15.4 & 36.3 & 45.2 & 50.1 & 43.8 & 34.5 \\
& ContextNet \cite{Kantorov2016ContextLocNet} &
\textbf{57.1} & 52.0 & 31.5 &  7.6 & 11.5 & 55.0 & 53.1 & 34.1 &  1.7 & 33.1 &
\textbf{49.2} & \textbf{42.0} & 47.3 & 56.6 & 15.3 & 12.8 & 24.8 & \textbf{48.9} & 44.4 & 47.8 & 36.3 \\
& WCCN \cite{Diba2017WCCN} &
43.9 & \textbf{57.6} & \textbf{34.9} & \textbf{21.3} & 14.7 & \textbf{64.7} & 52.8 & 34.2 &  6.5 & 41.2 &
20.5 & 33.8 & 47.6 & 56.8 & 12.7 & 18.8 & \textbf{39.6} & 46.9 & \textbf{52.9} & 45.1 & 37.3 \\
& OICR \cite{Tang2017OICR} &
53.1 & 57.1 & 32.4 & 12.3 & 15.8 & 58.2 & 56.7 & \textbf{39.6} &  0.9 & 44.8 &
39.9 & 31.0 & \textbf{54.0} & \textbf{62.4} &  4.5 & 20.6 & 39.2 & 38.1 & 48.9 & 48.6 & 37.9 \\
\cline{2-23}
& MELM &
56.4 & 54.7 & 30.9 & 21.1 & \textbf{17.3} & 52.8 & \textbf{60.0} & 36.1 &  3.9 & \textbf{47.8} &
35.5 & 28.9 & 30.9 & 61.0 &  5.8 & 22.8 & 38.8 & 39.6 & 42.1 & \textbf{54.8} & \textbf{38.4} \\
\hlinew{1.4pt}
\multirow{8}{*}{VGG16}& WSDDN \cite{Bilen2016Weakly} &
39.4 & 50.1 & 31.5 & 16.3 & 12.6 & 64.5 & 42.8 & 42.6 & 10.1 & 35.7 &
24.9 & 38.2 & 34.4 & 55.6 &  9.4 & 14.7 & 30.2 & 40.7 & 54.7 & 46.9 & 34.8 \\
& PDA \cite{Li2016Weakly} &
54.5 & 47.4 & \textbf{41.3} & 20.8 & \textbf{17.7} & 51.9 & 63.5 & 46.1 & 21.8 & 57.1 &
22.1 & 34.4 & 50.5 & 61.8 & 16.2 & 29.9 & 40.7 & 15.9 & 55.3 & 40.2 & 39.5 \\
& OICR \cite{Tang2017OICR} &
\textbf{58.0} & 62.4 & 31.1 & 19.4 & 13.0 & 65.1 & 62.2 & 28.4 & 24.8 & 44.7 &
30.6 & 25.3 & 37.8 & 65.5 & 15.7 & 24.1 & 41.7 & 46.9 & \textbf{64.3} & \textbf{62.6} & 41.2 \\
& Self-Taught \cite{Jie2017deepself} &
52.2 & 47.1 & 35.0 & 26.7 & 15.4 & 61.3 & 66.0 & \textbf{54.3} &  3.0 & 53.6 &
24.7 & 43.6 & 48.4 & 65.8 &  6.6 & 18.8 & 51.9 & 43.6 & 53.6 & 62.4 & 41.7 \\
& WCCN \cite{Diba2017WCCN} &
49.5 & 60.6 & 38.6 & \textbf{29.2} & 16.2 & \textbf{70.8} & 56.9 & 42.5 & 10.9 & 44.1 &
29.9 & 42.2 & 47.9 & 64.1 & 13.8 & 23.5 & 45.9 & 54.1 & 60.8 & 54.5 & 42.8 \\
& TS$^2$C  \cite{Wei2018TS2C} &
59.3 & 57.5 & 43.7 & 27.3 & 13.5 & 63.9 & 61.7 & 59.9 & 24.1 & 46.9 &
36.7 & 45.6 & 39.9 & 62.6 & 10.3 & 23.6 & 41.7 & 52.4 & 58.7 & 56.6 & 44.3 \\
& WeakRPN  \cite{Tang2018WeakRPN} &
57.9 & 70.5 & 37.8 &  5.7 & 21.0 & 66.1 & 69.2 & 59.4 &  3.4 & 57.1 &
57.3 & 35.2 & 64.2 & 68.6 & 32.8 & 28.6 & 50.8 & 49.5 & 41.1 & 30.0 & 45.3 \\

\cline{2-23}
& MELM &
55.6 & \textbf{66.9} & 34.2 & 29.1 & 16.4 & 68.8 & \textbf{68.1} & 43.0 & \textbf{25.0} & \textbf{65.6} &
\textbf{45.3} & \textbf{53.2} & 49.6 & \textbf{68.6} &  2.0 & 25.4 & \textbf{52.5} & \textbf{56.8} & 62.1 & 57.1 & \textbf{47.3} \\

\hlinew{1.4pt}
\multirow{2}{*}{\tabincell{c}{$Ens.$}}
& OICR-Ens. \cite{Tang2017OICR} &
58.5 & 63.0 & 35.1 & 16.9 & 17.4 & 63.2 & 60.8 & 34.4 &  8.2 & 49.7 &
41.0 & 31.3 & 51.9 & 64.8 & \textbf{13.6} & 23.1 & 41.6 & 48.4 & 58.9 & 58.7 & 42.0 \\
\cline{2-23}
& MELM-Ens. &
\textbf{60.3} & \textbf{65.0} & \textbf{39.5} & \textbf{29.0} & \textbf{17.5} & \textbf{66.1} & \textbf{66.4} & \textbf{44.8} & \textbf{18.6} & \textbf{59.0} &
\textbf{48.4} & \textbf{53.2} & \textbf{53.0} & \textbf{67.2} &         11.0  & \textbf{26.5} & \textbf{50.0} & \textbf{55.7} & \textbf{63.1} & \textbf{62.4} & \textbf{47.8} \\
\hlinew{1.4pt}
\end{tabular}
\end{table*}

\textbf{Min-entropy models.}
We denoted the min-entropy models by ``MELM-D'' and ``MELM-L'' in Table \ref{table_comp_ourself}, which respectively corresponded to object clique discovery and object localization. We trained the models by simply cascading the object clique discovery and object localization branches, without using the recurrent learning.
Table\ \ref{table_comp_ourself} shows that with VGGF we achieved 33.6\% and 36.0\% mAP for object clique discovery and object localization branches, which improved the baseline ``MELM-base'' by 2.1\% and 5.5\%. For VGG16, ``MELM-L'' significantly improved the ``MELM-base+Clique'' from 29.5\% to 40.1\%, with a 10.6\% margin at most. This fully demonstrated that the min-entropy models and their implementation with object clique discovery and object localization branches were pillars of our approach.

\textbf{Recurrent learning.}
In Table \ \ref{table_comp_ourself}, the recurrent learning algorithms ``MELM-D+RL'' and ``MELM-L+RL'', respectively achieved 34.5\% and 42.6\% mAP, improving the ``MELM-L'' (without recurrent learning) by 0.5\% and 2.4\%.  When using VGG16, ``MELM-D+RL'' and ``MELM-L+RL'' respectively achieved 34.5\% and 42.6\% mAP, improving the ``MELM-L'' by 1.9\% and 2.5\%.
These improvements showed that with recurrent learning, Fig.\ \ref{fig-melm-network}, the object clique discovery and object localization branches benefited from each other and thus were mutually enforced.

\textbf{Accumulated recurrent learning.} The models with accumulated recurrent learning were denoted by ``MELM-D+ARL'', ``MELM-L1+ARL'', and ``MELM-L2+ARL'' in Table \ \ref{table_comp_ourself}. In the learning procedure, the high scored proposals were accumulated into the next branch. When using two object localization branches, ``MELM-L2-ARL'' significantly improved the mAP of ``MELM-L-RL'' from 42.6\% to 46.4\% (+3.8\%). It further improved the mAP from 46.4\% to 47.3\% (+0.9\%) when using three branches, but did not significantly improve when using four.

\begin{table}[!t]
\renewcommand{\arraystretch}{1.3}
\newcommand{\tabincell}[2]{\begin{tabular}{@{}#1@{}}#2\end{tabular}}
\caption{Correct localization rate (\%) on the PASCAL VOC 2007 $trainval$ set. Comparison of MELM to the state-of-the-arts.}
\label{table_comp_loc_state_of_the_art}
\centering
\scriptsize
\begin{tabular}{@{}p{2.0cm}<{\centering}@{}|p{3cm}<{\centering}|@{}p{1.5cm}<{\centering}@{}}
\hlinew{1.4pt}
CNN & Method & mAP  \\
\hline
\multirow{8}{*}{\tabincell{c}{VGGF/\\AlexNet}}
& MILinear \cite{Ren2016Weakly}                 & 43.9 \\
& LCL+Context \cite{wang2015LCL}                & 48.5 \\
& PDA \cite{Li2016Weakly}                       & 49.8 \\
& WCCN \cite{Diba2017WCCN}                      & 52.6 \\
& Multi-fold MIL \cite{cinbis2015multi-fold-J}  & 54.2 \\
& WSDDN \cite{Bilen2016Weakly}                  & 54.2 \\
& ContextNet \cite{Kantorov2016ContextLocNet}   & 55.1 \\
\cline{2-3}
& MELM                                          & \textbf{58.4} \\
\hlinew{1.4pt}
\multirow{4}{*}{VGG16}
& PDA \cite{Li2016Weakly}                       & 52.4 \\
& WSDDN \cite{Bilen2016Weakly}                  & 53.5 \\
& WCCN \cite{Diba2017WCCN}                      & 56.7 \\
\cline{2-3}
& MELM                                          & \textbf{61.4} \\

\hlinew{1.4pt}
\end{tabular}
\end{table}

\begin{table}[!t]
\renewcommand{\arraystretch}{1.3}
\newcommand{\tabincell}[2]{\begin{tabular}{@{}#1@{}}#2\end{tabular}}
\caption{Detection mean average precision (\%) on the PASCAL VOC 2010, 2012, and the ILSVRC 2013 datasets. Comparison of MELM to the state-of-the-arts.}
\label{table_comp_state_101213}
\centering
\scriptsize
\begin{tabular}{p{0.9cm}<{\centering}|p{0.9cm}<{\centering}|p{2.0cm}<{\centering}|p{1.9cm}<{\centering}|p{0.8cm}<{\centering}}
\hlinew{1.4pt}
Dataset  & CNN & Method & Dataset Splitting & mAP  \\
\hline
\multirow{8}{*}{\tabincell{c}{PASCAL\\VOC\\2010}}    & \multirow{4}{*}{\tabincell{c}{VGGF/\\AlexNet}}& PDA  \cite{Li2016Weakly}         & train/val         & 21.4 \\
                             &                          & WCCN \cite{Diba2017WCCN}                              & trainval/test     & 28.8 \\
\cline{3-5}
                             &                          & MELM                                             & train/val         & \textbf{35.6} \\
                             &                          & MELM                                             & trainval/test     & \textbf{36.3} \\
\cline{2-5}
                             & \multirow{4}{*}{VGG16}   & PDA  \cite{Li2016Weakly}                              & train/val         & 30.7 \\
                             &                          & WCCN \cite{Diba2017WCCN}                              & trainval/test     & 39.5 \\
\cline{3-5}
                             &                          & MELM                                             & train/val         & \textbf{37.1} \\
                             &                          & MELM                                             & trainval/test     & \textbf{39.9} \\

\hlinew{1.4pt}
\multirow{15}{*}{\tabincell{c}{PASCAL\\VOC\\2012}}   & \multirow{7}{*}{\tabincell{c}{VGGF/\\AlexNet}}
                             & PDA \cite{Li2016Weakly}                                                          & train/val         & 22.4 \\&
                             & MILinear  \cite{Ren2016Weakly}                                                   & train/val         & 23.8 \\
                             &
                             & WCCN \cite{Diba2017WCCN}                                                         & trainval/test     & 28.4 \\
                             &                          & ContextNet \cite{Kantorov2016ContextLocNet}           & trainval/test     & 35.3 \\
                             &                          & OICR-VGGM  \cite{Tang2017OICR}                        & trainval/test     & 34.6 \\
\cline{3-5}
                             &                          & MELM                                             & train/val         & \textbf{36.2} \\
                             &                          & MELM                                             & trainval/test     & \textbf{36.4} \\
\cline{2-5}
                             & \multirow{8}{*}{VGG16}   & PDA  \cite{Li2016Weakly}                                                & train/val         & 29.1 \\
                             &                          & Self-Taught  \cite{Jie2017deepself}                   & train/val         & 39.0 \\
                             &                          & WCCN  \cite{Diba2017WCCN}                             & trainval/test     & 37.9 \\

                             &                          & OICR \cite{Tang2017OICR}                              & trainval/test     & 37.9 \\
                             &                          & Self-Taught \cite{Jie2017deepself}                    & trainval/test     & 38.3 \\
                             &                          & TS$^2$C \cite{Wei2018TS2C}                            & trainval/test     & 40.0 \\
\cline{3-5}
                             &                          & MELM                                             & train/val         & \textbf{40.2} \\
                             &                          & MELM                                             & trainval/test     & \textbf{42.4} \\

\hlinew{1.4pt}

\multirow{4}{*}{\tabincell{c}{ILSVRC\\2013}} & \multirow{4}{*}{\tabincell{c}{VGGF/\\AlexNet}}& MILinear  \cite{Ren2016Weakly}   & -                 &  9.6 \\
                             &                          & PDA  \cite{Li2016Weakly}                      & val1/val2         &  7.7 \\
                             &                          & WCCN  \cite{Diba2017WCCN}                             & -                 &  9.8 \\
\cline{3-5}
                             &                          & MELM                                             & val1/val2         & \textbf{13.4} \\
\cline{2-5}
\hlinew{1.4pt}
\end{tabular}
\end{table}

\begin{table}[!t]
\renewcommand{\arraystretch}{1.3}
\newcommand{\tabincell}[2]{\begin{tabular}{@{}#1@{}}#2\end{tabular}}
\caption{{\color{black}Detection and localization performance (\%) on MSCOCO 2014. Comparison of MELM to the state-of-the-arts.}}
\label{table_coco_2014}
\centering
\scriptsize
\color{black}
\begin{tabular}{
p{1.8cm}<{\centering}|
p{0.5cm}<{\centering}|
p{0.53cm}<{\centering}|
p{0.5cm}<{\centering}|
p{0.5cm}<{\centering}|
p{0.54cm}<{\centering}|
p{0.5cm}<{\centering}|
p{0.5cm}<{\centering}
}
\hlinew{1.4pt}
\multicolumn{8}{c}{Image Classification} \\
\hline
Method & mAP & F1-C & P-C & R-C & F1-O & P-O & R-O  \\
\hline
CAM \cite{zhou2016CAM} & 54.4 & - & - & - & - & - & - \\
\hline
SPN \cite{zhu2017soft} & 56.0 & - & - & - & - & - & - \\
\hline
ResNet-101 \cite{he2016deep} & 75.2 & 69.5 & 80.8 & 63.4 & 74.4 & 82.2 & 68.0 \\
\hline
MELM-VGG16 & 79.1 & 72.0 & 79.3 & 68.6 & 76.8 & 82.5 & 71.9 \\
\end{tabular}

\begin{tabular}{
p{0.7cm}<{\centering}|
p{1.55cm}<{\centering}|
p{1.2cm}<{\centering}|
p{1.05cm}<{\centering}|
p{1cm}<{\centering}|
p{0.8cm}<{\centering}
}
\hlinew{1.4pt}
\multicolumn{6}{c}{Pointing Localization (with class prediction)} \\
\hline
Method & WeakSup \cite{oquab2015object} & Pronet \cite{sun2016pronet} & DFM \cite{bency2016weakly} & SPN \cite{zhu2017soft} & MELM  \\
\hline
mAP & 41.2 & 43.5 & 49.2 & 55.3 & 65.1 \\
\end{tabular}

\begin{tabular}{
p{1.79cm}<{\centering}|
p{1.78cm}<{\centering}|
p{1.78cm}<{\centering}|
p{1.78cm}<{\centering}
}
\hlinew{1.4pt}
\multicolumn{4}{c}{Object Detection} \\
\hline
Method & CNN & mAP@.5 & mAP@[.5,.95]   \\
\hline
WSDDN \cite{Bilen2016Weakly} & VGGF & 10.1 & 3.1  \\
\hline
\multirow{2}{*}{MELM} & VGGF & 11.9 & 4.1  \\
\cline{2-4}
 & VGG16 & 18.8 & 7.8  \\
\hlinew{1.4pt}
\end{tabular}

\end{table}

\subsection{Performance and Comparison}

{\color{black}\subsubsection{PASCAL VOC datasets}}

\textbf{Weakly Supervised Object Detection.} Table \ \ref{table_comp_state_of_the_art} compared the detection performance of MELM with the state-of-the-art approaches on the PASCAL VOC 2007 dataset. It can be seen that MELM respectively achieved 38.4\% and 47.3\% with the VGGF and VGG16 models. With the popular VGG16 model, MELM respectively outperformed the OICR \cite{Tang2017OICR}, Self-Taught \cite{Jie2017deepself}, WCCN \cite{Diba2017WCCN}, WeakRPN \cite{Tang2018WeakRPN}, and TS$^2$C \cite{Wei2018TS2C} by 6.1\% (47.3\% vs. 41.2\%), 5.6\% (47.3\% vs. 41.7\%), 4.5\% (47.3\% vs. 42.8\%), 3.0\% (47.3\% vs. 44.3\%) and 2.0\% (47.3\% vs. 45.3\%), which were significant margins in terms of the challenging WSOD task. MELM using multiple networks (MELM-Ens.) outperformed OICR-Ens. (47.8\% mAP vs. 42.0\% mAP). {\color{black}To further improve the detection performance, we re-trained a Fast-RCNN detector using learned pseudo objects and a ResNet-101 network, and achieved 49.0\% mAP.}

Table \ \ref{table_comp_state_101213} compared the detection performance of MELM with the state-of the-art approaches on the VOC 2010 and VOC 2012 datasets. It can be seen that MELM usually outperformed the state-of-the-art approaches. On the VOC 2010 dataset, MELM with VGGF significantly outperformed WCCN \cite{Diba2017WCCN} by 7.5\% (36.3\% vs. 28.8\%)  with a VGGF model, and was comparable to it with a VGG16 model. On the VOC2012 dataset, with a VGGF model, MELM respectively outperformed WCCN \cite{Diba2017WCCN} and OICR \cite{Tang2017OICR} by 8.0\% ( 36.4\% vs. 28.4\%) and 1.8\% (36.4\% vs. 34.6\%). With a VGG16 model, MELM respectively outperformed WCCN \cite{Diba2017WCCN}, Self-Taught \cite{Jie2017deepself}, OICR \cite{Tang2017OICR}, and TS$^2$C \cite{Wei2018TS2C} by 4.5\% (42.4\% vs. 37.9\%), 4.1\% (42.4\% vs. 38.3\%), 4.5\% (42.4\% vs. 37.9\%) and 2.4\% (42.4\% vs. 40.0\%).

\begin{figure*}[!t]
    \begin{center}
       \subfloat[PASCAL VOC 2012]{\includegraphics[width=1\linewidth]{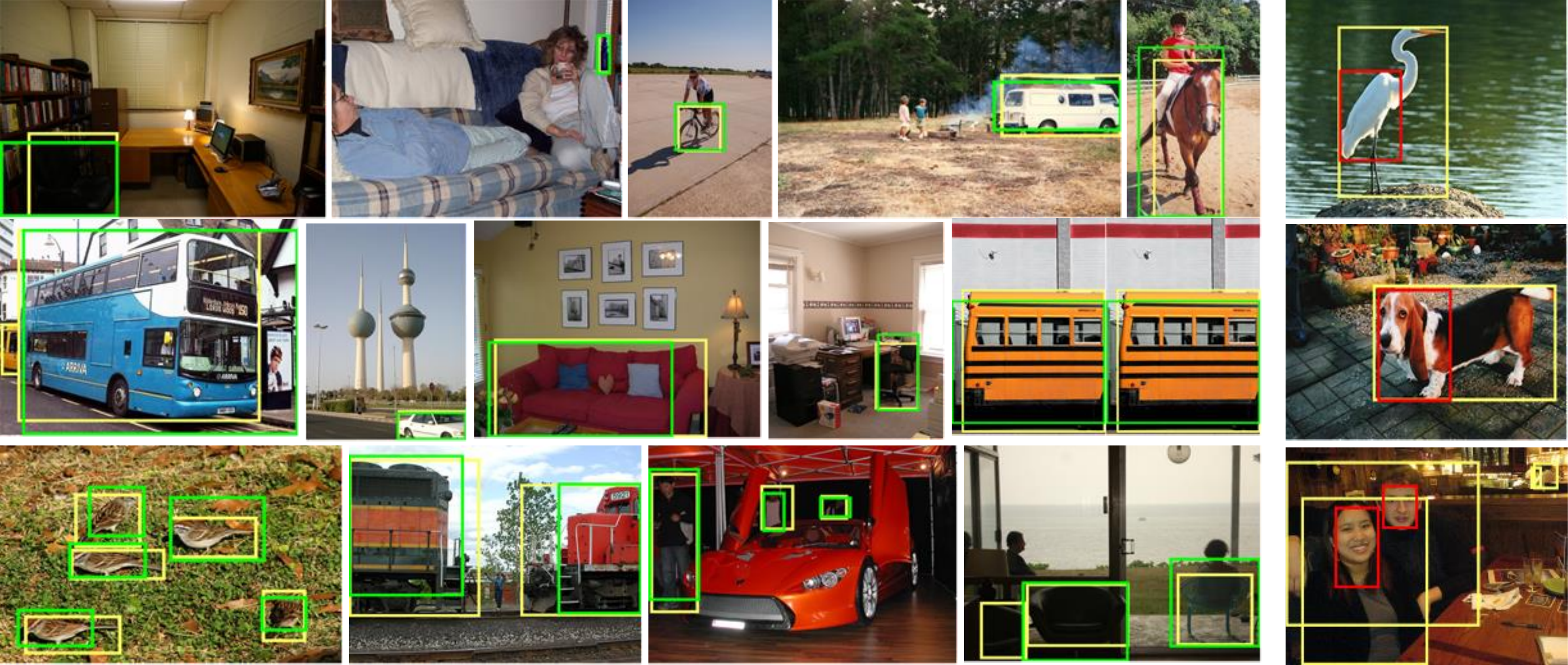}} \\
       \subfloat[MSCOCO 2014]{\includegraphics[width=1\linewidth]{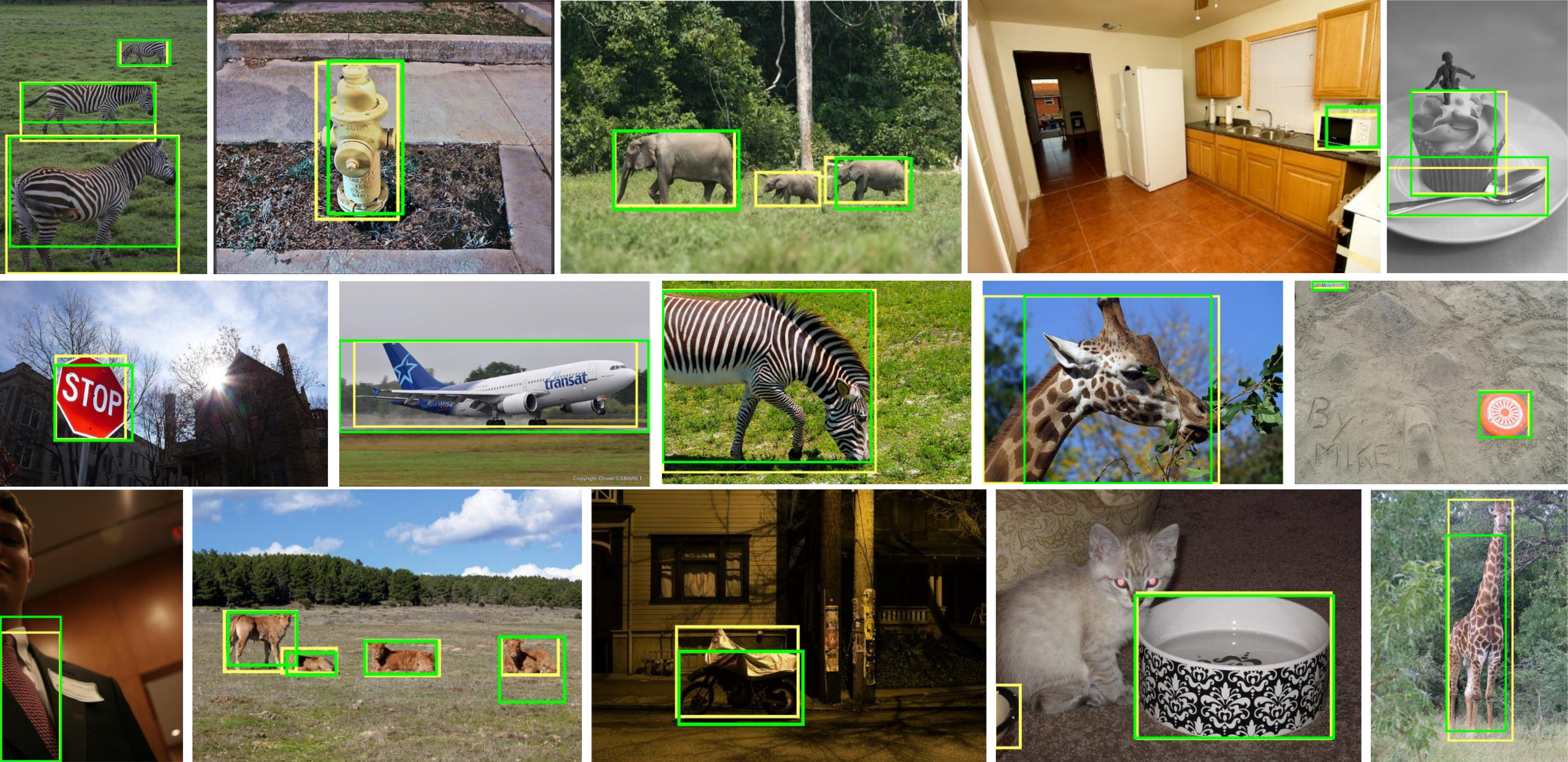}}
    \end{center}
    \caption{ {\color{black}Object detection examples on the PASCAL VOC 2012 and MS COCO 2014 datasets. Yellow bounding boxes denote ground-truth annotations, green boxes correct detection results and red boxes false detection results. (Best viewed in color).}}
    \label{detection_result}
\end{figure*}

{\color{black} Specifically, the detection performance for ``bicycle'' (+4.5\%), ``cow'' (+8.5\%), ``dining-table'' (+14.7\%), ``dog'' (+9.6\%) significantly improved, which shows the general effectiveness of MELM��

Despite of the average good performance, our approach failed on the ``person'' class, as shown in the last image of Fig.\ \ref{detection_result}(a).  ``Person'' is one of the most challenging class��
as people often involve great appearance variance from clothes, poses, and occlusions. Furthermore, the definition for ??person?? is not consistent. A ``person'' could be defined as a pedestrian, a head-and-shoulder, or just a human face. Given such ambiguous definition, what the algorithm can do is to localize the most discriminative part of a ``person'', e.g., the face. We also note that although the performance of ``person'' decreased, the average performance for all class significantly increased.

For the object classes with large appearance variance, we observed that the algorithm correctly classified the object regions but often failed to precisely localize them, i.e., the IoU between the learned bounding boxes and the groundtruth is smaller than 0.5. When using the ``pointing localization'' metric \cite{zhu2017soft}, the ``person'' class achieved 97.1\% localization accuracy, which shows potential to practical applications.}

Fig.\ \ref{detection_result} shows some of the detection examples. It can be seen that MELM precisely localize objects from clutter background and correctly localized multiple object regions in a single image.

\textbf{Weakly Supervised Object Localization.} The Correct Localization (CorLoc) metric \cite{deselaers2012Loc} was employed to evaluate the localization accuracy. CorLoc is the percentage of images for which the region of highest object score has at least 0.5 interaction-over-union (IoU) with the ground-truth object region. This experiment was done on the $trainval$ set because the region selection exclusively worked in the training process.

It can be seen in Table \ref{table_comp_loc_state_of_the_art} that with VGGF model, the mean CorLoc of MELM respectively outperformed the state-of-the-art WSDDN \cite{Bilen2016Weakly} and WCCN \cite{Diba2017WCCN} by 4.2\% (58.4\% vs. 54.2\%) and 5.8\% (58.4\% vs. 52.6\%).  With the VGG16 model, it respectively outperformed the state-of-the-art WSDDN \cite{Bilen2016Weakly} and WCCN \cite{Diba2017WCCN} by 7.9\% (61.4\% vs. 53.5\%) and 4.7\% (61.4\% vs. 56.7\%).
Noticeably, on the ``bus'', ``car'', ``chair'', and ``table'' classes, MELM outperformed the compared state-of-the-art methods up to 7$\sim$15\%. This shows that the clique-based min-entropy strategy is more effective than the image segmentation strategy used in WCCN.

\begin{table}[!t]
\renewcommand{\arraystretch}{1.3}
\newcommand{\tabincell}[2]{\begin{tabular}{@{}#1@{}}#2\end{tabular}}
\caption{\color{black} Image classification mAP (\%) on the PASCAL VOC 2007 $test$ set. Comparison of MELM to the state-of-the-arts.}
\label{table_comp_cls_state_of_the_art}
\centering
\scriptsize
\begin{tabular}{@{}p{2.0cm}<{\centering}@{}|p{3cm}<{\centering}|@{}p{1.5cm}<{\centering}@{}}
\hlinew{1.4pt}
CNN & Method & mAP  \\
\hline
\multirow{5}{*}{\tabincell{c}{VGGF/\\AlexNet}}
& MILinear \cite{Ren2016Weakly}         & 72.0 \\
& AlexNet \cite{krizhevsky2012imagenet} & 82.4 \\
& WSDDN \cite{Bilen2016Weakly}          & 85.3 \\
& WCCN \cite{Diba2017WCCN}              & \textbf{87.8} \\
\cline{2-3}
&MELM                                   & \textbf{87.8} \\
\hlinew{1.4pt}
\multirow{4}{*}{VGG16}
& VGG16 \cite{Simonyan2014Very}         & 89.3 \\
& WSDDN \cite{Bilen2016Weakly}          & 89.7 \\
& WCCN \cite{Diba2017WCCN}              & 90.9 \\
\cline{2-3}
&MELM                                   & \textbf{93.1} \\
\hlinew{1.4pt}
\end{tabular}
\end{table}

\textbf{Image Classification.}
The object clique discovery and object localization components highlighted informative regions and suppressed disturbing backgrounds, which also benefited image classification.
As shown in Tab.\ \ref{table_comp_cls_state_of_the_art}, with the VGGF model, MELM achieved 87.8\% mAP. With the VGG16 model, MELM achieved 93.1\% mAP, which respectively outperformed WSDDN \cite{Bilen2016Weakly} and WCCN \cite{Diba2017WCCN} up to 3.4\% (93.1\% vs. 89.7\%) and 2.2\% (93.1\% vs. 90.9\%). It is noteworthy that MELM outperformed the VGG16 network, specifically trained for image classification, by 3.8\% mAP (93.1\% vs. 89.3\%).

\subsubsection{Large-scale datasets}
On the ILSVRC2013 dataset with 200 object classes, Table \ \ref{table_comp_state_101213}, MELM with VGGF outperformed the WCCN approach by 3.6\% (13.4\% vs. 9.8\%). On the MS COCO 2014 dataset, we evaluated the image classification, pointing localization, and object detection performance and compared it with the state-of-the-arts. The evaluation metrics for image classification included macro/micro precision (P-C and P-O), macro/micro recall (R-C and R-O), macro/micro F1-measure (F1-C and F1-O) \cite{wu2016unified}. It can be seen in Table. \ref{table_coco_2014} that for image classification MELM outperformed SPN \cite{zhu2017soft} by 23.1\% (79.1\% vs. 56\%). For pointing localization, MELM outperformed SPN by 9.8\% (65.1\% vs. 55.3\%). For object detection, MELM outperformed WSDDN.
With these experiments, we set new baselines for weakly supervised object detection on large-scale datasets.

\section{Conclusion}

In this paper, we proposed an effective deep min-entropy latent model (MELM) for weakly supervised object detection (WSOD). {\color{black}MELM was deployed as three components of clique partition, object clique discovery, and object localization,} and was unified with the deep learning framework in an {\color{black} integrated} manner. {\color{black} By partitioning and discovering cliques,} MELM provided a new way to learn latent object regions from redundant object proposals. With the min-entropy principle, it can principally reduce the variance of positive instances and alleviate the ambiguity of detectors.  {\color{black} With the recurrent learning algorithm,} MELM improved the performance of weakly supervised detection, weakly supervised localization, and image classification, in striking contrast with state-of-the-art approaches. {\color{black}The underlying reality is that min-entropy results in minimum randomness of an information system and the recurrent learning takes advantages of continuation optimization, which provides fresh insights for weakly supervised learning problems.}

\section*{Appendix}

For succinct representation, we denote ${E_{\left( {{\cal X,Y}} \right)}}\left( {{{\cal H}_c},\theta } \right)$, ${E_{\left( {{\cal X,Y},{{\cal H}_c}} \right)}}\left( {h,\theta } \right)$, ${L_{\left( {{\cal X,Y}} \right)}}\left( {{{\cal H}_c},\theta } \right)$, and ${L_{\left( {{\cal X,Y},{{\cal H}_c}} \right)}}\left( {h,\theta } \right)$ as ${E}\left( {{{\cal H}_c},\theta } \right)$, $E (h,\theta )$, $L\left( {{{\cal H}_c},\theta } \right)$, and $L\left( {h,\theta } \right)$, respectively.

\textbf{Derivation for object clique discovery.} Given the object score $s\left( {{y},{h};\theta } \right)$ as the input of the entropy models, its gradient can be computed as
\begin{equation}\label{grad_Ld}
    \begin{split}
        &\frac{{\partial L({{\cal H}_c},\theta )}}{{\partial s\left( {y,h;\theta } \right)}} = \sum\limits_{y',h'} {\frac{{\partial L({{\cal H}_c},\theta )}}{{\partial p\left( {y',h';\theta } \right)}}\frac{{\partial p\left( {y',h';\theta } \right)}}{{\partial s\left( {y,h;\theta } \right)}}}  \\
        &= \sum\limits_{y',h'} {\left( {y'\frac{{\partial E\left( {{{\cal H}_c},\theta } \right)}}{{\partial p\left( {y',h';\theta } \right)}} + \frac{{y' - 1}}{{1 - p\left( {y',h';\theta } \right)}}} \right)\frac{{\partial p\left( {y',h';\theta } \right)}}{{\partial s\left( {y,h;\theta } \right)}}},
    \end{split}
\end{equation}
where the partial derivation of ${E}\left( {{\cal H}_c,\theta } \right)$ with respect to $p\left( {y,h;\theta } \right)$ is computed as
\begin{equation}\label{grad_Ed}
    \begin{split}
    &\frac{{\partial {E}\left( {{\cal H}_c,\theta } \right)}}{{\partial p\left( {{y'},{h'};\theta } \right)}} = \frac{{ - 1}}{{\sum\limits_c {{w_{{{\cal H}_c}}}\sum\limits_{h \in {{\cal H}_c}} {p\left( {y,h;\theta } \right)} } }} \cdot \\
    &\left( {{\frac{1}{{\left| {{\cal H}_c^{'}} \right|}}\left( {\sum\limits_{h \in {\cal H}_c^{'}} {{p\left( {y',h;\theta } \right)}} } \right)\left( {\sum\limits_{y \ne {y'}} {p\left( {y,h;\theta } \right)} } \right)} }\right.\\
    &\left. {{/{{\left( {\sum\limits_y {p\left( {y,h;\theta } \right)} } \right)}^2} + {w_{{\cal H}_c^{'}}}} }\right),
    \end{split}
\end{equation}
where ${\cal{H}_c}^{'}$ is the clique including $h'$. The partial derivation of $p\left( {y,h;\theta } \right)$ with respect to $s\left( {{y},h;\theta } \right)$ is computed as
\begin{equation}\label{grad_pyh}
    \begin{split}
    \frac{{\partial p\left( {{y'},{h'};\theta } \right)}}{{\partial s\left( {{y},h;\theta } \right)}} = \left\{ {\begin{array}{*{20}{c}}
    { - s\left( {{y'},h';\theta } \right)s\left( {{y},h;\theta } \right), h\neq h' \;or\; y\neq y' ,}\\
    {s\left( {{y'},h';\theta } \right) - s{{\left( {{y},h;\theta } \right)}^2}, otherwise.}
    \end{array}} \right.
    \end{split}
\end{equation}
\textbf{Derivation for object localization.} In Eq.\ (\ref{loss_object_localization}), the term ${{w_h}p\left( {y,h;\theta} \right)}$ is used as a pseudo label for $h$, which does not back-propagate gradients. Therefore, the derivation for object localization can be simply computed as
\begin{equation}\label{grad_Ll}
    \begin{split}
        \frac{{\partial {L}(h,\theta)}}{{\partial s\left( {{y},h;\theta} \right)}} &= \sum\limits_{{y'}, {h'}} {\frac{{\partial {L}(h,\theta)}}{{\partial p\left( {{y'},{h'};\theta } \right)}}\frac{{\partial p\left( {{y'},{h'};\theta } \right)}}{{\partial s\left( {{y},h;\theta } \right)}}} \\
        &= \sum\limits_{{{y'}, {h'} \in \left\{ {{\cal H}_1^*,{\cal H}_2^*,...} \right\}}} {{w_{h'}}\frac{{\partial p\left( {y',h';\theta } \right)}}{{\partial s\left( {{y},h;\theta } \right)}}}.
    \end{split}
\end{equation}
The partial derivation of $L(h,\theta)$ with respect to $s\left( {{y},h;\theta } \right)$ is calculated with Eq.\ (\ref{grad_pyh}) and Eq.\ (\ref{grad_Ll}).

\ifCLASSOPTIONcompsoc
\section*{Acknowledgments}
This work was supported in part by the NSFC under Grant 61836012, 61671427, and 61771447, and Beijing Municipal Science and Technology Commission under Grant Z181100008918014. Qixiang Ye is the corresponding author.

\else
  \section*{Acknowledgment}
\fi

\ifCLASSOPTIONcaptionsoff
  \newpage
\fi

\bibliographystyle{IEEEtran}
\bibliography{IEEEabrv,melm}

\vfill
\begin{IEEEbiography}[{\includegraphics[width=1in,height=1.25in,clip,keepaspectratio]{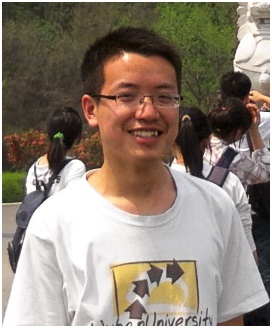}}]
{Fang Wan} received the B.S. degree from Wuhan University, Wuhan, China, in 2013. Since 2013, he has been a Ph.D student in the School of Electronic, Electrical and Communication Engineering, University of Chinese Academy of Sciences, Beijing, China. His research interests include computer vision and machine learning, specifically for weakly supervised learning and visual object detection.
\end{IEEEbiography}

\begin{IEEEbiography}[{\includegraphics[width=1in,height=1.25in,clip,keepaspectratio]{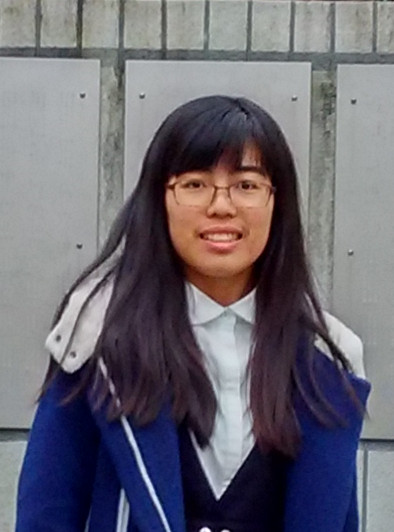}}]
{Pengxu Wei} received the B.S. degree in computer science from the China University of Mining and Technology, Beijing, China, in 2011, and the Ph.D. degree from University of Chinese Academy of Sciences in 2018. Since 2018, she has been  a research scientist at Sun Yat-sen University, Guangzhou, China. Her research interests include computer vision and machine learning, specifically for data-driven vision and scene image recognition. \end{IEEEbiography}

\begin{IEEEbiography}[{\includegraphics[width=1in,height=1.25in,clip,keepaspectratio]{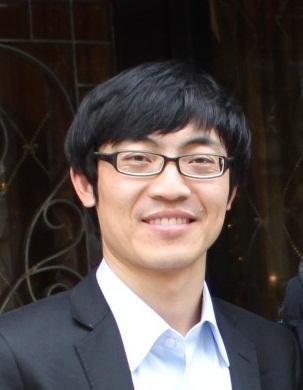}}]
{Zhenjun Han} (M'12) received the B.S. degree from Tianjin University, Tianjin, China,
in 2006 and the M.S. and Ph.D. degrees from University of Chinese Academy of Sciences (UCAS), Beijing,
China, in 2009 and 2012, respectively. Since 2013, he has been an Associate Professor of UCAS.
His research interests include visual object detection, tracking and recognition. He has published about 40 papers in refereed conferences and journals. \end{IEEEbiography}

\begin{IEEEbiography}[{\includegraphics[width=1in,height=1.25in,clip,keepaspectratio]{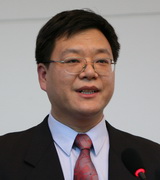}}]
{Jianbin Jiao} (M'10) received the B.S., M.S., and Ph.D. degrees from the Harbin Institute of Technology (HIT), China, in 1989, 1992, and 1995, respectively. From 1997 to 2005, he was an Associate Professor with HIT. Since 2006, he has been a Professor with the University of the Chinese Academy of Sciences, Beijing, China. In the research areas about image processing and pattern recognition. He has authored over 50 papers in refereed conferences and journals.
\end{IEEEbiography}

\begin{IEEEbiography}[{\includegraphics[width=1in,height=1.25in,clip,keepaspectratio]{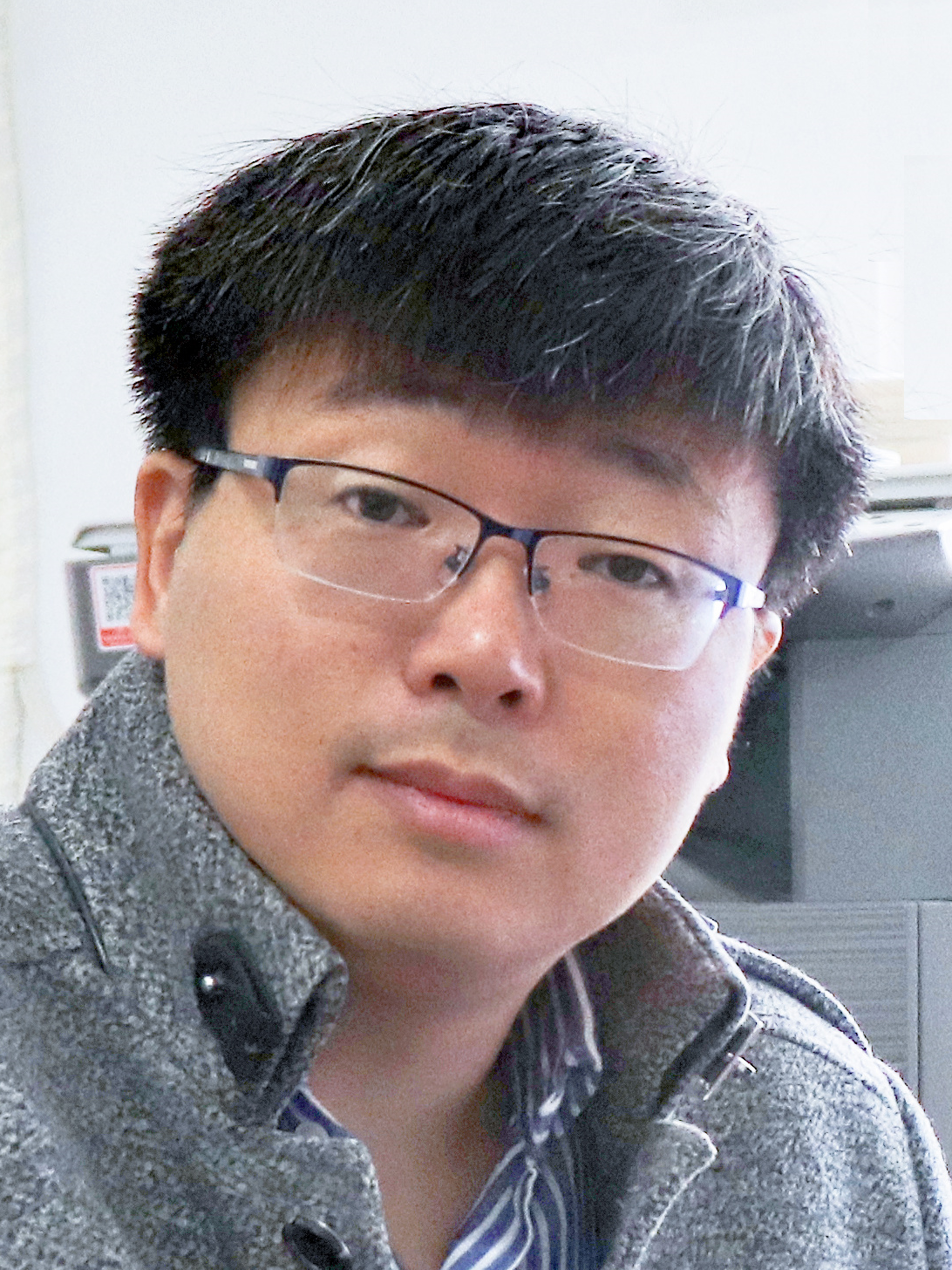}}]{Qixiang Ye}
(M'10-SM'15) received the B.S. and M.S. degrees from Harbin Institute of Technology, China, in 1999 and 2001, respectively, and the Ph.D. degree from the Institute of Computing Technology, Chinese Academy of Sciences in 2006. He has been a professor with the University of Chinese Academy of Sciences (UCAS) since 2009, and was a visiting assistant professor with the University of Maryland, College Park until 2013. His research interests include visual object detection and machine learning. He has published more than 80 papers in refereed conferences and journals including IEEE CVPR, ICCV, ECCV, and PAMI, and received the Sony Outstanding Paper Award.
\end{IEEEbiography}
\vfill

\end{document}